\documentclass[10pt,twocolumn,letterpaper]{article} 
\PassOptionsToPackage{table}{xcolor} 
\usepackage[pagenumbers]{cvpr} 

\usepackage{amsmath,amsfonts,bm}

\def\eqref#1{equation~\ref{#1}}

\def\1{\bm{1}}

\DeclareMathAlphabet{\mathsfit}{\encodingdefault}{\sfdefault}{m}{sl}
\SetMathAlphabet{\mathsfit}{bold}{\encodingdefault}{\sfdefault}{bx}{n}

\usepackage[utf8]{inputenc}
\usepackage[T1]{fontenc}
\usepackage{url}
\usepackage{booktabs}
\usepackage{amsfonts}
\usepackage{amsmath}
\usepackage{amssymb}
\usepackage{nicefrac}
\usepackage{microtype}
\usepackage{graphicx}
\usepackage{multirow}
\usepackage{enumitem}
\usepackage{array}
\usepackage{adjustbox}
\usepackage{needspace}
\usepackage{float}
\usepackage{placeins} 
\usepackage{makecell}
\usepackage[most]{tcolorbox}
\tcbuselibrary{listings,breakable}
\usepackage{listings}
\usepackage{algorithm}
\usepackage{algpseudocode}
\usepackage{etoc}
\definecolor{cvprblue}{rgb}{0.21,0.49,0.74}
\usepackage[breaklinks,colorlinks,allcolors=cvprblue]{hyperref}

\algnewcommand\LineComment[1]{\State \(\triangleright\) #1}
\lstdefinestyle{promptstyle}{
  basicstyle=\ttfamily\footnotesize,
  breaklines=true,
  breakatwhitespace=false,
  keepspaces=true,
  columns=fullflexible,
  showstringspaces=false
}

\let\standardaddcontentsline\addcontentsline
\renewcommand{\addcontentsline}[3]{}

\title{ScenePilot: Plug-and-Play 3D Scene Repair across Generators}

\author{%
\textbf{Jiawei Zhang}$^{1}$ \quad
\textbf{Hongsong Wang}$^{1}$ \quad
\textbf{Pan Zhou}$^{2}$ \\
$^{1}$Southeast University \quad
$^{2}$Singapore Management University \\
\texttt{\{jiaweizhang,hongsongwang\}@seu.edu.cn \quad panzhou3@gmail.com}
}

\begin{document}
\maketitle
\begin{figure*}[t]
	\centering
	\includegraphics[width=0.96\linewidth]{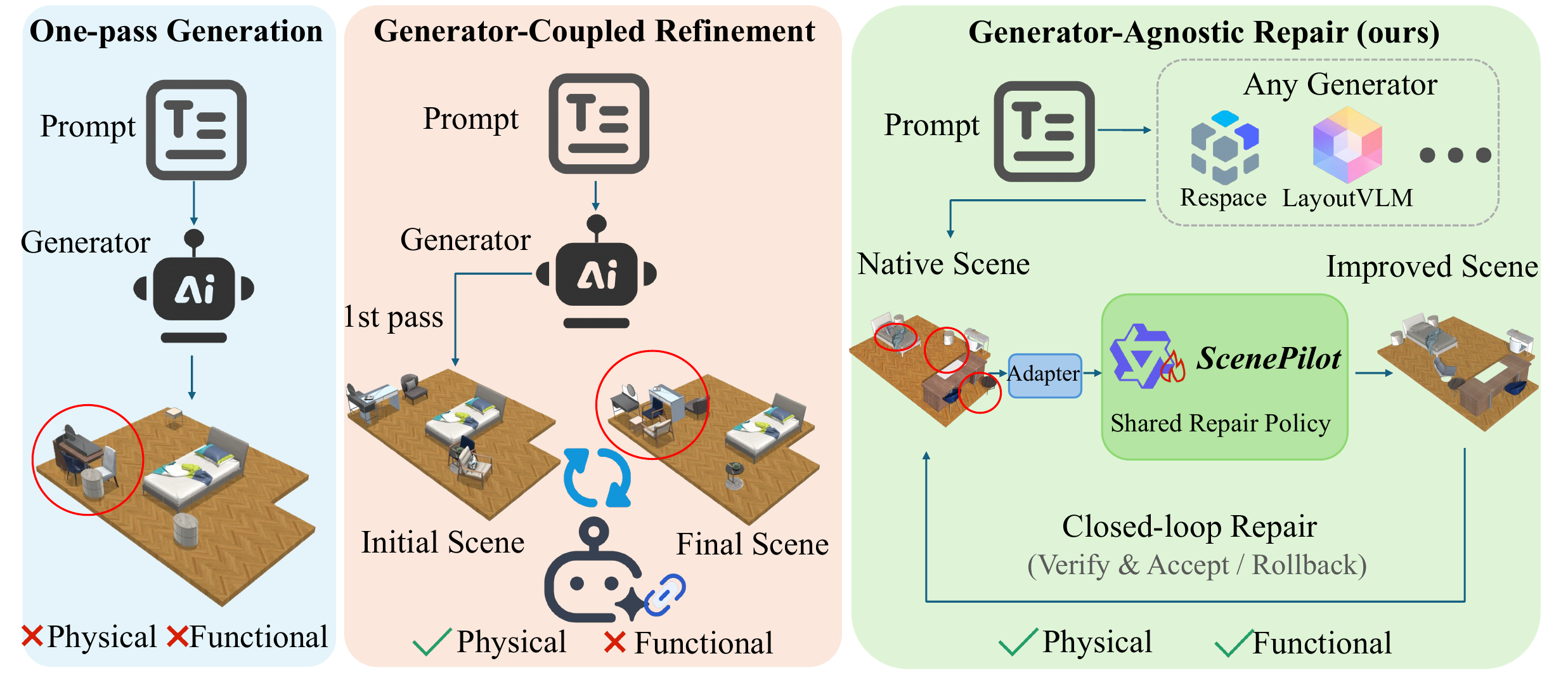}
	\vspace{-0.6em}
	\caption{\textbf{Comparison of scene generation and repair paradigms.}
        One-pass generation may leave residual geometric or functional errors, while generator-coupled refinement remains tied to a specific pipeline. \textbf{ScenePilot} instead treats repair as a generator-agnostic capability, applying a shared policy across heterogeneous backends through lightweight adapters and conservative closed-loop verification with accept-or-rollback decisions.}
	\label{fig:intro}
	\vspace{-1.2em}
\end{figure*}
\begin{abstract}
Text-driven 3D scene generators have made substantial progress, yet their outputs often retain collisions, boundary violations, orientation errors, and structural inconsistencies. Existing correction mechanisms are typically coupled to generator-specific scene representations and execution pipelines, making repair difficult to reuse across heterogeneous systems. We introduce \textbf{ScenePilot}, a plug-and-play 3D scene repair policy designed to make corrective behavior transferable across generators. ScenePilot decouples repair reasoning from generator-specific execution through a canonical scene-editing interface, allowing one shared multimodal policy to operate across heterogeneous generators without modifying or retraining their underlying generation models. At inference time, ScenePilot performs conservative closed-loop repair: each round proposes at most one atomic edit, executes it tentatively, and retains the change only when a shared verifier confirms improvement. We further construct \textbf{SceneReverse}, which transforms clean 3D scenes into executable recovery trajectories and provides generator-independent supervision for learning atomic repair decisions. Combined with supervised learning and reward-guided policy optimization based on realized improvements, SceneReverse enables repair learning without relying on failure traces from any target generator. On SceneEval, the same frozen ScenePilot policy consistently improves scenes from five heterogeneous generators unseen during policy training, without backend-specific policy fine-tuning, using only lightweight backend adapters. These results support transferable 3D scene repair as a reusable capability that can be attached to existing generators through a canonical interface.
\end{abstract}

\section{Introduction}
Text-driven 3D indoor scene generation turns a request such as ``a bedroom with a bed, two nightstands, and a desk'' into an editable 3D environment. This capability supports interior design, embodied AI, robotics simulation, and virtual-world authoring \citep{deitke2022procthor,xia2026sage}. Unlike image generation, however, the result must not merely look plausible: objects cannot intersect walls or one another, related furniture should be arranged coherently, and enough free space must remain for navigation and everyday use \citep{wang2019planit,lin2024instructscene}. A scene may therefore appear reasonable while still containing serious geometric or functional errors.

Early data-driven approaches learn indoor scene distributions through autoregressive modeling, transformers, or diffusion \citep{wang2018deeppriors,ritchie2019fastflex,wang2021sceneformer,paschalidou2021atiss,tang2024diffuscene}. Complementary approaches address scene geometry and appearance through radiance fields, room synthesis, and panoramic reconstruction \citep{bautista2022gaudi,song2023roomdreamer,zhou2025holodreamer,zhang2024berfscene}. Language-guided methods improve semantic control through layout planning, structured representations, and procedural generation \citep{feng2023layoutgpt,berdoz2025reason3d,celen2025idesign,ocal2024sceneteller,zhou2025scenex}. Meanwhile, spatial search and zone-based organization improve scene planning \citep{deng2025global,zhang2026zonemaestro}, while agentic methods incorporate visual feedback and iterative correction \citep{yang2025sceneweaver,ling2025scenethesis,zhao2026scenerevis}. 
As illustrated in Fig.~\ref{fig:intro}, existing paradigms either leave residual errors after one-pass generation or couple iterative refinement to a specific pipeline. In contrast, ScenePilot treats repair as a shared capability that operates across heterogeneous generators through a canonical adapter and conservative closed-loop verification. This motivates a distinct question: can corrective behavior be learned once and reused across heterogeneous generators?

This work studies the less explored problem of \emph{transferable 3D scene repair}: how can a single policy improve scenes from heterogeneous text-driven 3D generators without being tied to their internal representations or generation procedures?  
The goal is not merely to optimize another generator, but to make \emph{scene repair itself transferable}.  Existing generators differ substantially in scene formats, placement mechanisms, and supported editing operations, making repair difficult to transfer across systems. Moreover, generated scenes often contain residual collisions, boundary violations, orientation errors, and structural inconsistencies that require iterative correction. Addressing this problem introduces three key challenges. 1) \textbf{Can a single repair policy generalize across heterogeneous scene generators?} Existing optimization mechanisms are typically tailored to a specific scene representation and execution pipeline, and therefore cannot be directly reused when the underlying generator changes.   
2) \textbf{How can such a shared policy reliably repair imperfect scenes?} 
A common interface makes transfer possible, but not necessarily effective: generated scenes contain collisions, boundary violations, orientation errors, and structural inconsistencies, while each correction changes the scene and may introduce new defects. Reliable repair therefore requires closed-loop interaction---repeatedly observing the updated scene, selecting a targeted edit, assessing its realized effect, and deciding whether to continue or stop while preserving already valid regions.  
3) \textbf{How can this transferable behavior be learned without collecting supervision for every generator?} 
Training on generator-specific failure traces would undermine the goal of plug-and-play transfer, yet standard 3D scene datasets provide mostly clean final layouts rather than imperfect states paired with executable recovery decisions. 
These challenges call for a plug-and-play framework that decouples repair from individual generators, iteratively refines scenes from fresh multimodal observations, and learns corrective behavior from generator-independent supervision.

\textbf{Contributions.} 
To address these three challenges, we propose \textbf{ScenePilot}, a plug-and-play repair policy that turns scene refinement from a generator-specific component into a transferable capability. 
1) \textbf{Generator-Agnostic Repair Interface:}
ScenePilot separates \emph{what should be repaired} from \emph{how a particular generator executes the repair}. A lightweight adapter maps each generator's native scene into a shared object-centric representation, exposes its supported editing operations, and translates canonical actions---\texttt{translate}, \texttt{rotate}, \texttt{scale}, \texttt{add}, and \texttt{delete}---back into native edits. This allows one frozen policy to operate across heterogeneous generators without generator-specific policy fine-tuning. 
2) \textbf{Conservative Closed-Loop Repair:}
ScenePilot performs repair as an iterative perception--action loop rather than one-shot post-processing. At each round, the multimodal policy predicts one atomic edit or \texttt{STOP}; the candidate edit is executed on a temporary scene and retained only if a shared verifier confirms improvement, otherwise it is rolled back. The retained scene is then re-rendered for the next decision, enabling traceable multi-step correction while preserving already valid regions. 
3) \textbf{SceneReverse: Generator-Independent Process Supervision:} 
We construct \textbf{SceneReverse} from high-quality 3D-FRONT scenes \citep{fu2021front} by applying controlled perturbations and reversing them into recovery trajectories. Intermediate states provide atomic repair supervision, while recovered states provide explicit \texttt{STOP} targets. Combined with supervised learning and reward-guided policy optimization based on realized improvements, SceneReverse enables the policy to learn repair behavior without relying on failure traces from any particular generator.

\section{Related Work}

\noindent\textbf{3D Indoor Scene Generation and Layout Priors.}
Indoor scene synthesis spans learned object arrangements \citep{wang2018deeppriors,ritchie2019fastflex,wang2021sceneformer,paschalidou2021atiss,tang2024diffuscene,maillard2024debara}, graph-conditioned generation \citep{lin2024instructscene,chen2025graph2scene,yang2025mmgdreamer}, and language-driven layout prediction \citep{yang2024holodeck,berdoz2025reason3d,bucher2025respace,wan2026nala}. Complementary methods connect layouts with appearance synthesis \citep{fang2025ctrlroom}, use image intermediaries \citep{gu2025artiscene}, or address open-set geometry and pose estimation \citep{shi2025scenemaker}. These approaches primarily construct scenes with different representations and controls. ScenePilot instead focuses on transferable repair: retrieved priors support optional incremental generation, while the shared repair policy remains independent of the planner and generator.

\noindent\textbf{Scene Optimization and Repair.}
Existing methods address scene validity through human-motion conditioning \citep{yi2022mime}, relational placement constraints \citep{yang2024holodeck}, differentiable layout optimization \citep{sun2025layoutvlm}, denoising \citep{maillard2024debara}, physically grounded guidance \citep{yang2024physcene}, and differentiable simulation \citep{lin2026pat3d}. Agentic approaches incorporate visual refinement and self-reflection \citep{ling2025scenethesis,yang2025sceneweaver}. These mechanisms improve scene quality, but do not by themselves establish the transfer of a shared repair policy across heterogeneous generators. ScenePilot studies this setting by separating repair decisions from backend-specific execution through a canonical scene--action interface and lightweight adapters.

\noindent\textbf{Process Supervision and Policy Learning.}
More broadly, iterative refinement with self-feedback and interleaved reasoning and acting motivate closed-loop correction beyond a single generation step \citep{madaan2023selfrefine,yao2023react}. Recent methods introduce multi-turn reinforcement learning \citep{pan2025metaspatial}, reward-guided layout reasoning \citep{ran2025directlayout}, and supervised learning followed by policy alignment \citep{bucher2025respace,zhang2026zonemaestro}. Closely related, SceneReVis combines reverse-engineered construction trajectories, supervised fine-tuning, and multi-turn reinforcement learning for visual scene refinement \citep{zhao2026scenerevis}. ScenePilot shares trajectory supervision and policy optimization, but focuses on generator-agnostic repair: one policy checkpoint is evaluated on outputs from multiple generators unseen during policy training, without backend-specific policy fine-tuning. Recovery trajectories supervise executable interventions, and the same policy supports both completed and intermediate scene states.

\section{ScenePilot}
\label{sec:method}

ScenePilot is a plug-and-play intervention policy for heterogeneous 3D scene generators. Its central design principle is to decouple \emph{intervention reasoning} from \emph{generator-specific execution}: one shared multimodal policy operates on a canonical scene representation, while lightweight adapters normalize native states and translate canonical actions into executable edits. The same frozen policy can therefore be applied either to completed scenes for post-hoc refinement or to intermediate states during incremental generation, without generator-specific policy fine-tuning.

\begin{figure*}[t]
    \centering
    \includegraphics[width=0.96\linewidth]{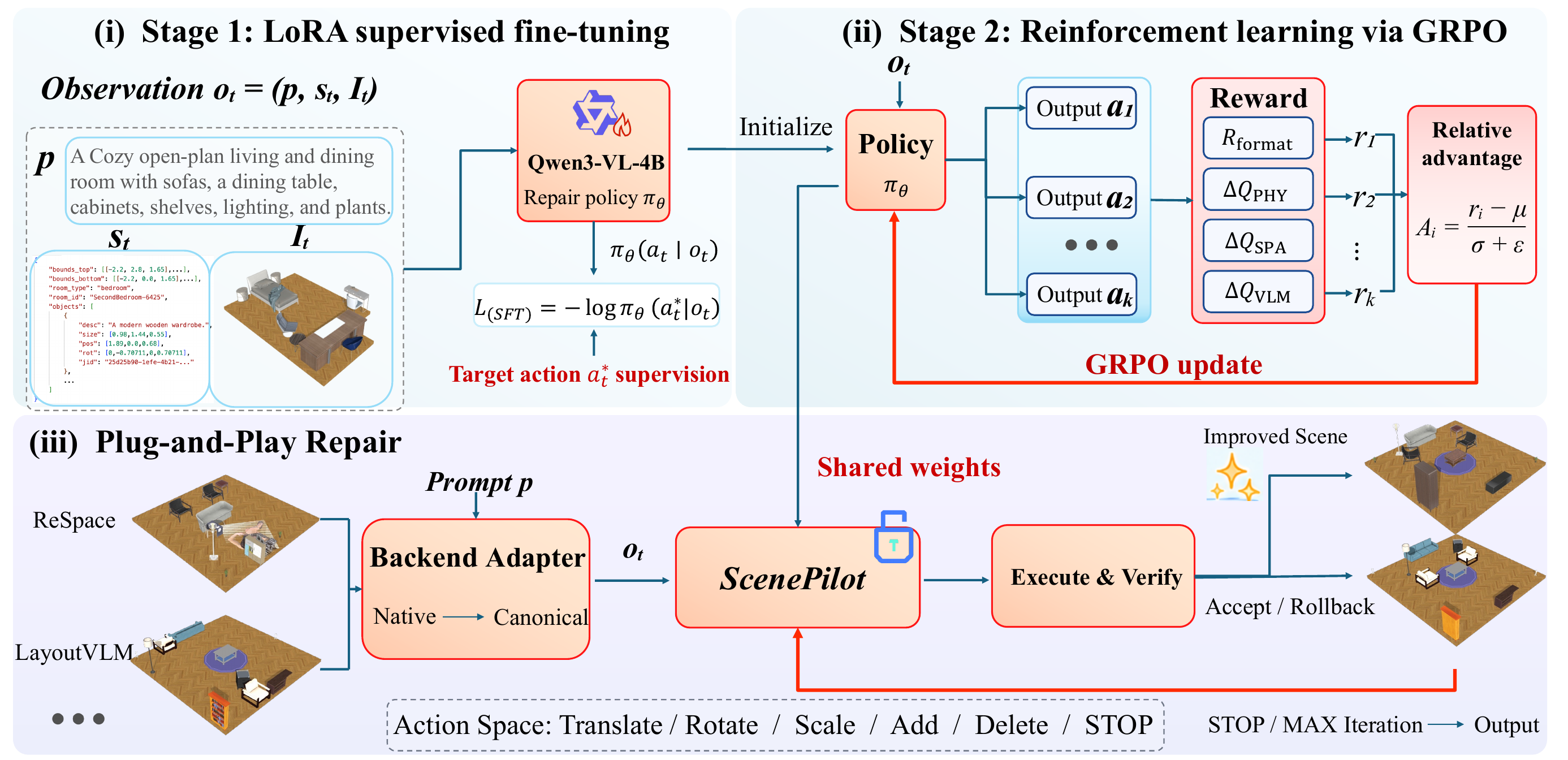}
    \caption{\textbf{Overview of ScenePilot.} (i) LoRA supervised fine-tuning adapts Qwen3-VL-4B-Instruct \citep{qwen2025qwen3vl} to predict repair actions from user prompts, structured scene states, and rendered views. (ii) GRPO further optimizes the policy using format validity and physical, spatial, and visual-semantic improvement rewards. (iii) Backend adapters map native scenes into a canonical representation, enabling plug-and-play repair across heterogeneous generators with one shared checkpoint. Candidate edits are executed and verified through accept-or-rollback decisions until STOP or the iteration limit is reached.}
    \label{fig:method_pipeline}
    \vspace{-0.5em}
\end{figure*}





\subsection{Generator-Agnostic Repair Interface}
\label{sec:interface}

Different 3D scene generators represent the same physical scene using different coordinate systems, object formats, and editing APIs. ScenePilot therefore does not require generators to share their internal representations. Instead, it defines a common object-centric repair interface that standardizes the \emph{semantics} of scene states and editing operations, while leaving generator-specific execution to lightweight
backend adapters.

\noindent\textbf{Shared scene state.}
Let $S_t^{(g)}$ denote the native scene produced by generator
$g\in\mathcal{G}$ at repair step $t$. Before being observed by the repair policy, the scene is mapped to a canonical state
\begin{equation}
    z_t
    =
    \left(
        B_t,
        \left\{
        (\mathrm{id}_i,c_i,\mathbf{p}_i,\theta_i,\mathbf{s}_i)
        \right\}_{i=1}^{N_t}
    \right),
    \label{eq:canonical_state}
\end{equation}
where $B_t$ denotes the room boundary and each object is represented by a stable identifier $\mathrm{id}_i$, semantic category $c_i$, metric position $\mathbf{p}_i$, yaw angle $\theta_i$, and physical size $\mathbf{s}_i$. Backend-specific units, coordinate axes, rotation conventions, and boundary formats are normalized into this shared metric representation. The purpose of this representation is not merely to unify serialization formats, but to provide consistent object references and geometric semantics on which a single repair policy can reason across heterogeneous generators.



\noindent\textbf{Canonical edits and capabilities.}
ScenePilot operates in a shared editing space $\mathcal{A}=\{\texttt{translate},\texttt{rotate},\texttt{scale},\texttt{delete},\texttt{add}\}$, with \texttt{STOP} defined separately as the terminal decision. These operations specify \emph{what} change should be made to the scene without assuming how a particular generator implements that change. Not every backend needs to support the full action space. Each generator therefore exposes a capability descriptor $C_g$, which defines its executable subset $\mathcal{A}_g \subseteq \mathcal{A}$ together with the valid parameter ranges of the supported operations. The capability descriptor is provided to the shared policy as part of its observation, allowing the same policy to reason under different execution constraints. Unsupported operations are not silently converted into another action; they are rejected by the capability-aware validation layer described in Sec.~\ref{sec:repair}.

\noindent\textbf{Adapter-based execution.}
For each generator, we implement a lightweight interface
\begin{equation}
    \mathcal{I}_g
    =
    (\phi_g,\psi_g,C_g),
    \label{eq:backend_interface}
\end{equation}
where $\phi_g$ maps the native scene $S_t^{(g)}$ to the canonical state $z_t$, $\psi_g$ grounds a validated canonical action into an executable native edit, and $C_g$ specifies the capabilities exposed by the backend. The adapter is restricted to representation conversion and action grounding:it does not decide which object should be repaired and contains no generator-specific repair policy.

Consequently, changing the underlying scene generator changes only $\phi_g$, $\psi_g$, and $C_g$, while the repair-policy parameters $\theta$ remain shared. This separation allows ScenePilot to determine \emph{what should be changed} independently of \emph{how that change is executed} by a particular generator, forming the basis for plug-and-play
transfer across heterogeneous scene-generation systems.

\subsection{Conservative Closed-Loop Repair}
\label{sec:repair}

A generator-agnostic interface makes it possible to execute a shared repair policy across heterogeneous backends, but an executable edit is not necessarily beneficial. A modification that resolves one local defect may introduce another, while an overly aggressive edit may unnecessarily alter an already plausible scene. ScenePilot therefore performs repair as a conservative closed loop in which each atomic proposal is executed on a temporary scene and its realized effect is assessed before the modification is committed.

\noindent\textbf{Multimodal atomic decisions.}
At repair step $t$, ScenePilot observes the current canonical scene together with two complementary rendered views and the capability descriptor of the current backend:
\begin{equation}
    o_t=(I_t,z_t,C_g),
    \qquad
    I_t=
    \left\{
        I_t^{\mathrm{diag}},
        I_t^{\mathrm{ann\text{-}top}}
    \right\},
    \label{eq:policy_observation}
\end{equation}
where the diagonal view captures object geometry and local spatial interactions, the annotated top-down view exposes the global layout and stable object references, and $z_t$ provides precise object poses and physical sizes. The capability descriptor $C_g$ specifies which canonical operations are available for the current backend.

Given the current observation, the shared policy predicts exactly one atomic decision:
\begin{equation}
    \hat a_t
    \sim
    \pi_\theta(\cdot\mid o_t),
    \qquad
    \hat a_t\in
    \mathcal{A}\cup\{\texttt{STOP}\},
    \label{eq:repair_policy}
\end{equation}
where $\mathcal{A}$ is the canonical edit space defined in Sec.~\ref{sec:interface}. Each non-terminal decision targets one object with one canonical operation. Restricting each round to a single edit makes the resulting state transition attributable to a specific intervention and directly matches the atomic supervision used for policy learning in Sec.~\ref{sec:learning}. If the policy predicts \texttt{STOP}, the repair process terminates without modifying the current scene.

\noindent\textbf{Execution and conservative acceptance.}
Before execution, each non-terminal proposal is passed through a capability-aware validation layer:
\begin{equation}
    a_t=
    \operatorname{ValidateAndProject}
    (\hat a_t;C_g,S_t^{(g)}),
    \label{eq:validate_project}
\end{equation}
which checks the output schema, object references, backend capabilities, and valid parameter ranges. Malformed, unsupported, or ungroundable proposals are rejected rather than silently converted into another operation.

A valid action is grounded through the backend adapter and first applied to a temporary copy of the native scene:
\begin{equation}
    \widetilde{S}_{t+1}^{(g)}
    =
    F_g\!\left(
        S_t^{(g)},
        \psi_g(a_t;S_t^{(g)})
    \right),
    \label{eq:tentative_execution}
\end{equation}
where $F_g$ denotes the native scene-editing operation of generator $g$. ScenePilot then measures the realized change using a shared scene-quality score
\begin{equation}
\begin{aligned}
    V(S)
    =
    \lambda_{\mathrm{phy}}Q_{\mathrm{PHY}}(S)
    &+
    \lambda_{\mathrm{spa}}Q_{\mathrm{SPA}}(S)\\
    &+
    \lambda_{\mathrm{vlm}}Q_{\mathrm{VLM}}(S),
\end{aligned}
    \label{eq:repair_score}
\end{equation}
where $Q_{\mathrm{PHY}}$, $Q_{\mathrm{SPA}}$, and $Q_{\mathrm{VLM}}$ measure physical validity, spatial consistency, and higher-level visual-semantic quality, respectively. The realized improvement is
\begin{equation}
    \Delta V_t
    =
    V(\widetilde{S}_{t+1}^{(g)})
    -
    V(S_t^{(g)}).
    \label{eq:repair_delta}
\end{equation}

The tentative edit is committed only when the resulting scene improves the shared score:
\begin{equation}
    S_{t+1}^{(g)}
    =
    \begin{cases}
        \widetilde{S}_{t+1}^{(g)}, & \Delta V_t>\epsilon,\\
        S_t^{(g)}, & \text{otherwise}.
    \end{cases}
    \label{eq:conservative_commit}
\end{equation}
Thus, capability-aware validation determines whether a proposed edit can be executed, while the acceptance rule determines whether its realized consequence should be retained. Because this decision is made from the resulting scene rather than from the textual action itself, the same closed-loop procedure can be applied across heterogeneous backends.

\noindent\textbf{Closed-loop continuation and termination.}
After either acceptance or rollback, ScenePilot reconstructs the canonical state and re-renders the retained scene before invoking the policy again. Consequently, each decision is conditioned on the scene that actually results from previous repair rounds rather than on a predicted or assumed next state. The loop continues until the policy predicts \texttt{STOP} or the maximum repair budget $T_{\max}$ is reached.

The three components therefore have distinct responsibilities: the shared policy determines \emph{what should be changed}, the backend interface determines whether and how the proposal \emph{can be executed}, and the acceptance mechanism determines whether the realized modification \emph{should be retained}. The complete inference algorithm, including capability validation, tentative execution, accept-or-rollback decisions, and termination, is provided in Appendix.

\subsection{Learning Shared Repair Policy}
\label{sec:learning}

Training a transferable repair policy requires supervision that describes how an imperfect scene can be improved, rather than only how a final scene should look. However, existing 3D scene datasets mainly provide clean layouts and do not contain intermediate degraded states paired with executable corrective decisions. Collecting failure trajectories from each target generator would also compromise generator-agnostic transfer. We therefore construct \textbf{SceneReverse}, a generator-independent repair trajectory dataset derived from high-quality 3D-FRONT scenes.

\begin{figure*}[t]
    \centering
    \includegraphics[width=0.96\linewidth]{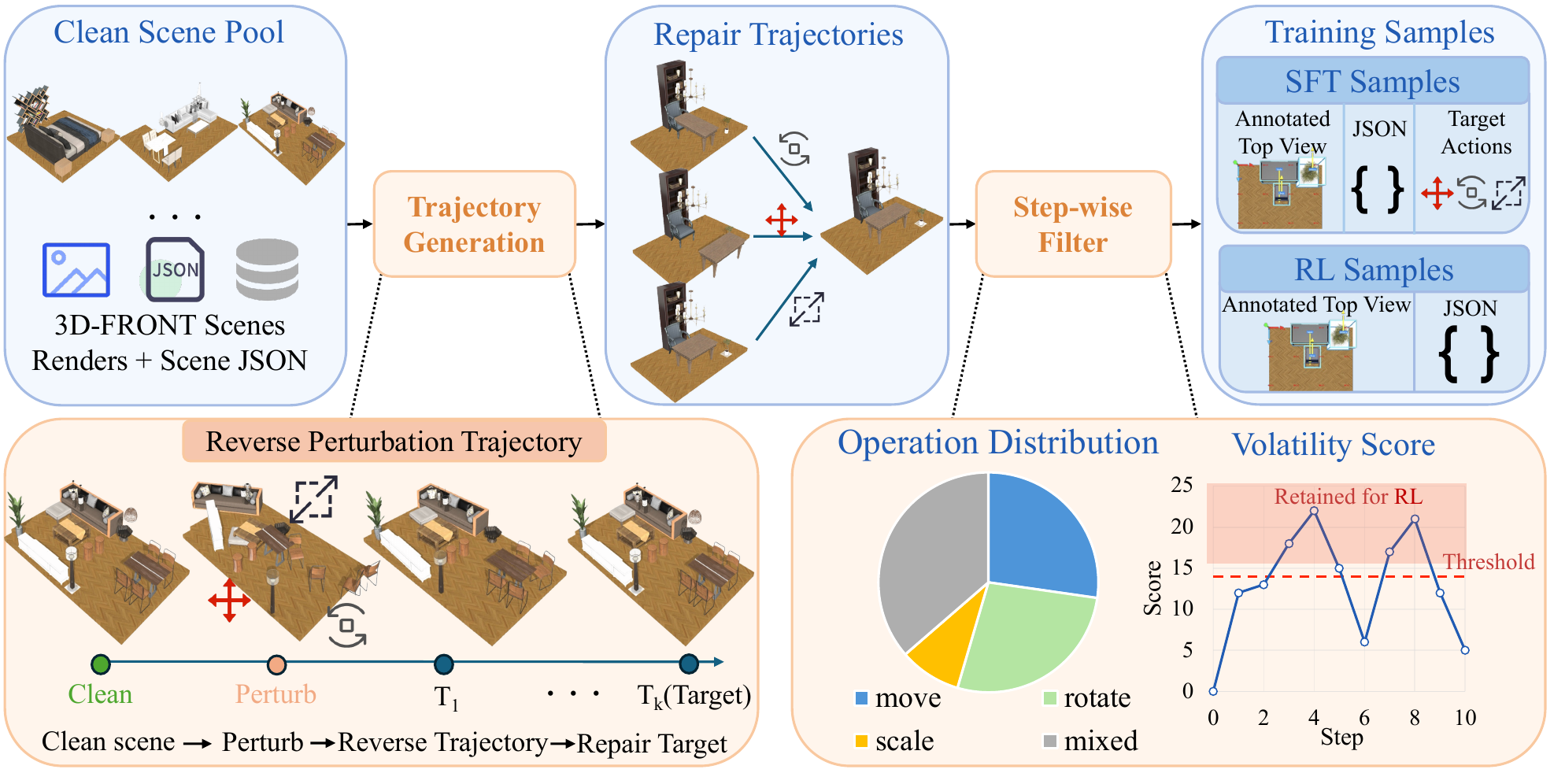}
    \caption{\textbf{Construction of SceneReverse.} Starting from a clean 3D-FRONT scene, we apply executable structured perturbations in the canonical edit space to obtain progressively degraded states. Reversing these perturbations produces atomic recovery decisions, while recovered terminal states provide explicit \texttt{STOP} supervision.}
    \label{fig:scenereverse_pipeline}
    \vspace{-0.5em}
\end{figure*}



\noindent\textbf{Executable recovery trajectories.}
Given a clean scene $S^\star$, SceneReverse generates degraded intermediate states through reversible object-level perturbations defined in the same canonical edit space as the repair policy. By reversing these perturbations, we obtain executable recovery trajectories consisting of atomic state--action transitions. Each transition records a degraded scene state and the corresponding recovery action, while the recovered terminal state is paired with a \texttt{STOP} decision.

Importantly, the recovery action induced by the reverse process is not assumed to be the unique optimal repair. Indoor scene repair is inherently multi-modal: different edits may lead to physically and functionally valid outcomes. SceneReverse therefore provides one executable and verifiable recovery path rather than a single deterministic ground truth.

Each recovery step is converted into an atomic state--action pair $(o_t^{\mathrm{r}},a_t^\star)$ following the same observation structure used during inference. Specifically, $o_t^{\mathrm{r}}$ contains the current rendered views, canonical scene state, and a capability descriptor indicating the available canonical operations. The recovered terminal state is paired with \texttt{STOP}, teaching the policy that a satisfactory scene should be preserved rather than modified unnecessarily. The final SceneReverse pool contains $9{,}582$ unique rooms, $20{,}012$ validated recovery trajectories, and $53{,}599$ atomic recovery steps. Detailed filtering criteria, difficulty coverage, data splits, and action statistics are provided in the Appendix.

\noindent\textbf{Supervised process learning.}
We first use SceneReverse to teach the policy grounded atomic repair behavior through supervised fine-tuning. Given a multimodal observation $o_t$, the policy is trained to predict either the demonstrated recovery action $a_t^\star$ or the terminal \texttt{STOP} decision:
\begin{equation}
    \mathcal{L}_{\mathrm{SFT}}
    =
    -
    \mathbb{E}_{(o_t,a_t^\star)\sim\mathcal{D}_{\mathrm{SR}}}
    \left[
        \log \pi_\theta(a_t^\star\mid o_t)
    \right],
    \label{eq:sft_objective}
\end{equation}
where $\mathcal{D}_{\mathrm{SR}}$ denotes the SceneReverse training set. Since each sample contains exactly one atomic target decision, the supervision directly matches the one-action-per-round inference protocol in Sec.~\ref{sec:repair}. This stage teaches multimodal scene understanding, object grounding, canonical action prediction, structured-output compliance, and termination on recovered states.

To prepare the shared policy for heterogeneous backend capabilities, we additionally expose it to different capability configurations during SFT. Partial-capability examples disable selected non-target operations while keeping the demonstrated target action executable, so the policy learns to condition its decision on the currently available action set rather than assuming that every backend supports the full canonical interface. Generator identity itself is never provided as an input. Consequently, the policy is trained to associate repair decisions with scene content and available operations rather than with a particular generator.

\noindent\textbf{Learning from realized improvements.}
Supervised process learning provides a grounded initialization, but the demonstrated inverse action is only one valid recovery decision and need not be the only beneficial edit for a given state. We therefore further optimize the SFT policy according to the realized consequences of its sampled actions. For each repair state, the current policy samples a group of candidate atomic decisions. Each candidate is parsed and validated through the same canonical interface used during inference, and executable non-terminal actions are applied to the current scene to obtain temporary next states.

For each candidate, we define the reward as
\begin{equation}
\begin{aligned}
    R
    =
    \lambda_{\mathrm{fmt}}R_{\mathrm{format}}
    &+
    \lambda_{\mathrm{phy}}\Delta Q_{\mathrm{PHY}} \\
    &+
    \lambda_{\mathrm{spa}}\Delta Q_{\mathrm{SPA}}
    +
    \lambda_{\mathrm{vlm}}\Delta Q_{\mathrm{VLM}},
\end{aligned}
    \label{eq:grpo_reward}
\end{equation}
where $R_{\mathrm{format}}$ measures whether the predicted decision follows the canonical output schema and can be grounded into an executable command. The remaining terms measure the scene-quality changes directly caused by the sampled action:
\begin{equation}
    \Delta Q_m
    =
    Q_m(\widetilde{S}_{t+1})
    -
    Q_m(S_t),
    \label{eq:reward_delta}
\end{equation}
where $m\in\{\mathrm{PHY},\mathrm{SPA},\mathrm{VLM}\}$, $Q_{\mathrm{PHY}}$ measures physical validity, $Q_{\mathrm{SPA}}$ measures spatial and relational consistency, and $Q_{\mathrm{VLM}}$ provides visual-semantic feedback from the rendered scene. Using improvement rather than absolute scene quality assigns credit to the consequence of the intervention and allows alternative actions to receive positive feedback even when they differ from the demonstrated inverse action.

We optimize these candidate decisions with Group Relative Policy Optimization (GRPO) \citep{shao2024grpo}, which compares sampled actions within the same repair state and regularizes the updated policy toward the SFT reference policy. The standard group-relative advantage, clipping objective, KL regularization, and complete optimization hyperparameters are provided in the appendix. Importantly, policy optimization is performed at the level of individual repair decisions rather than complete multi-round trajectories, providing localized credit assignment to the scene transition directly induced by each action.

At inference time, these learned atomic decisions are composed through the conservative closed-loop procedure in Sec.~\ref{sec:repair}. After each retained modification, the scene is re-canonicalized and re-rendered before the next policy decision. SceneReverse provides generator-independent process supervision for learning \emph{how to repair}, while realized-improvement optimization allows the policy to move beyond imitation of a single synthetic recovery path.

\section{Experiments}
\label{sec:experiments}

\begin{table*}[t]
\centering
\caption{\textbf{Evaluation on the SceneEval.} Parentheses in the Overall column denote relative percentage improvements over the corresponding Base row.
All official SceneEval submetrics are reported under the two benchmark groups, Fidelity and Plausibility.
The Overall is computed from the nine metrics, and $\mathrm{COL}_{\mathrm{sc}}$ is reported for completeness but excluded from Overall to avoid double-counting collision.}
\label{tab:main_cross_generator}
\small
\setlength{\tabcolsep}{2.4pt}
\resizebox{\linewidth}{!}{%
\begin{tabular}{llccccccccccc}
\toprule
& &
\multicolumn{4}{c}{SceneEval Fidelity}
& \multicolumn{6}{c}{SceneEval Plausibility}
& \multirow{2}{*}{Overall$\uparrow$} \\
\cmidrule(lr){3-6}
\cmidrule(lr){7-12}
Generator & Method
& CNT$\uparrow$
& ATR$\uparrow$
& OOR$\uparrow$
& OAR$\uparrow$
& COL$_{\mathrm{ob}}\downarrow$
& COL$_{\mathrm{sc}}\downarrow$
& SUP$\uparrow$
& NAV$\uparrow$
& ACC$\uparrow$
& OOB$\downarrow$
& \\
\midrule
\multirow{2}{*}{ReSpace}
& Base
& 80.18 & 49.47 & 17.89 & 18.24
& 31.44 & 55.56 & 12.07 & 100.00 & 61.12 & 5.78
& 52.77 \\
& + ScenePilot
& 83.47 & 49.73 & 36.76 & 34.60
& 18.28 & 37.78 & 10.84 & 100.00 & 59.67 & 2.66
& 59.79 \textbf{(+13.3\%)} \\

\midrule
\multirow{2}{*}{LayoutVLM}
& Base
& 41.19 & 22.26 & 8.60 & 23.29
& 36.09 & 69.00 & 67.96 & 98.75 & 85.91 & 4.14
& 44.34 \\
& + ScenePilot
& 41.19 & 39.53 & 16.90 & 24.71
& 33.21 & 54.00 & 68.03 & 97.88 & 87.42 & 2.89
& 50.52 \textbf{(+13.9\%)} \\

\midrule
\multirow{2}{*}{Reason-3D}
& Base
& 62.19 & 44.31 & 25.98 & 33.76
& 14.43 & 28.28 & 40.70 & 99.95 & 75.65 & 13.85
& 56.79 \\
& + ScenePilot
& 65.32 & 47.63 & 32.88 & 34.10
& 12.86 & 25.91 & 38.47 & 99.87 & 73.12 & 9.64
& 59.15 \textbf{(+4.2\%)} \\

\midrule
\multirow{2}{*}{LayoutGPT}
& Base
& 19.54 & 18.98 & 2.87 & 5.24
& 12.96 & 30.00 & 28.24 & 100.00 & 47.29 & 73.11
& 25.98 \\
& + ScenePilot
& 28.61 & 23.77 & 7.63 & 6.32
& 8.31 & 27.00 & 29.04 & 100.00 & 63.81 & 23.61
& 34.60 \textbf{(+33.2\%)} \\

\midrule
\multirow{2}{*}{InstructScene}
& Base
& 23.39 & 27.01 & 9.01 & 6.06
& 36.32 & 83.84 & 28.56 & 99.55 & 76.92 & 16.10
& 33.97 \\
& + ScenePilot
& 27.49 & 30.91 & 18.41 & 23.63
& 31.03 & 77.24 & 28.56 & 99.62 & 75.81 & 13.07
& 42.51 \textbf{(+25.1\%)} \\
\bottomrule
\end{tabular}}
\vspace{-0.4em}
\end{table*}

\begin{table*}[t]
\centering
\caption{\textbf{Repair-policy comparison.} Parentheses in the Overall column denote relative percentage improvements over Qwen3-VL-4B.
All policies are evaluated on the same frozen ReSpace generator outputs using the full official SceneEval submetrics.
Overall is computed from the same nine core metrics as in Table~\ref{tab:main_cross_generator}.}
\label{tab:policy_comparison}
\small
\setlength{\tabcolsep}{2.3pt}
\resizebox{\linewidth}{!}{%
\begin{tabular}{lccccccccccc}
\toprule
&
\multicolumn{4}{c}{SceneEval Fidelity}
& \multicolumn{6}{c}{SceneEval Plausibility}
& \multirow{2}{*}{Overall$\uparrow$} \\
\cmidrule(lr){2-5}
\cmidrule(lr){6-11}
Policy
& CNT$\uparrow$
& ATR$\uparrow$
& OOR$\uparrow$
& OAR$\uparrow$
& COL$_{\mathrm{ob}}\downarrow$
& COL$_{\mathrm{sc}}\downarrow$
& SUP$\uparrow$
& NAV$\uparrow$
& ACC$\uparrow$
& OOB$\downarrow$
& \\
\midrule

Qwen3-VL-4B
& 79.12 & 48.83 & 6.30 & 24.54
& 29.25 & 47.14 & 15.41 & 100.00 & 62.95 & 3.97
& 52.35 \\

ScenePilot-SFT
& 82.11 & 51.89 & 29.87 & 27.66
& 22.00 & 44.98 & 18.77 & 99.75 & 55.66 & 3.33
& 57.80 \textbf{(+10.4\%)} \\

ScenePilot-SFT+GRPO
& 83.47 & 49.73 & 36.76 & 34.60
& 18.28 & 37.78 & 10.84 & 100.00 & 59.67 & 2.66
& 59.79 \textbf{(+14.2\%)} \\

w/o verify
& 78.74 & 47.69 & 27.91 & 25.70
& 21.70 & 42.13 & 17.42 & 100.00 & 53.45 & 3.92
& 55.75 \textbf{(+6.5\%)} \\

\midrule

Expert-GPT-5.4$^\dagger$
& 83.22 & 59.62 & 43.71 & 47.64
& 15.40 & 18.74 & 23.69 & 100.00 & 69.17 & 1.69
& 66.33 \textbf{(+26.7\%)} \\
\bottomrule
\end{tabular}}
\vspace{-0.4em}
\end{table*}

\begin{table*}[t]
\centering
\caption{\textbf{Average perceptual scores from two independent VLM judges.}
Scores range from 1 to 10, with higher values indicating better perceptual quality. GPT-5.4 and Gemini-3-Flash independently evaluate each scene on layout coherence, relation fidelity, and object completeness without observing method identities or paired before/after results.}
\label{tab:vlm_score}
\small
\setlength{\tabcolsep}{3.8pt}
\resizebox{\linewidth}{!}{%
\begin{tabular}{lcccccccc}
\toprule
& \multicolumn{4}{c}{GPT-5.4}
& \multicolumn{4}{c}{Gemini-3-Flash} \\
\cmidrule(lr){2-5} \cmidrule(lr){6-9}
Policy
& Layout$\uparrow$
& Relation$\uparrow$
& Completeness$\uparrow$
& Average$\uparrow$
& Layout$\uparrow$
& Relation$\uparrow$
& Completeness$\uparrow$
& Average$\uparrow$ \\
\midrule

ReSpace
& 7.10 & 6.57 & 7.05 & 6.91
& 6.33 & 6.24 & 7.05 & 6.54 \\

w/ ScenePilot (ours)
& \textbf{7.24} & \textbf{7.19} & \textbf{7.57} & \textbf{7.33}
& \textbf{6.81} & 6.86 & \textbf{7.48} & \textbf{7.05} \\

LayoutVLM
& 6.29 & 6.10 & 6.33 & 6.24
& 6.14 & 6.19 & 6.24 & 6.19 \\

w/ ScenePilot (ours)
& 6.73 & 6.82 & 6.35 & 6.63
& 6.57 & \textbf{7.12} & 6.27 & 6.65 \\

\bottomrule
\end{tabular}}
\vspace{-0.4em}
\end{table*}

\begin{figure*}[t]
\centering
\includegraphics[width=\linewidth]{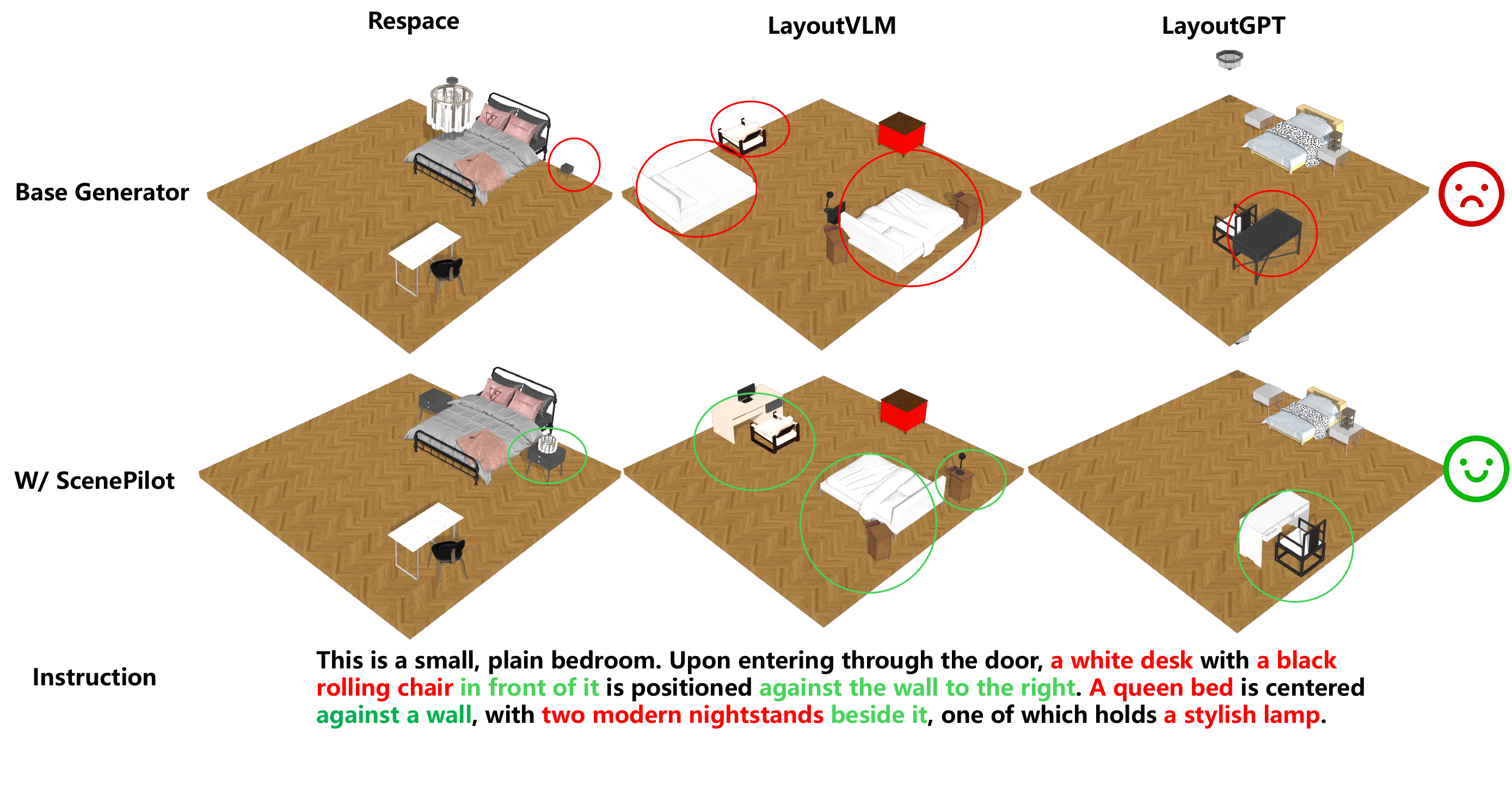}
\vspace{-0.4em}
\caption{\textbf{Qualitative examples of plug-and-play repair across generators.} Each column compares the native generator output with the result after applying the same frozen ScenePilot policy under the same instruction. Highlighted regions illustrate representative local repairs without regenerating the scene
from scratch.}
\label{fig:qualitative_compare}
\vspace{-0.5em}
\end{figure*}

We evaluate ScenePilot around four questions. 
\textbf{RQ1:} Can one shared repair policy improve scenes produced by heterogeneous 3D generators without backend-specific policy tuning?
\textbf{RQ2:} Does the learned policy transfer to a held-out generator through only the canonical adapter and capability interface?
\textbf{RQ3:} How much do process supervision, GRPO, capability conditioning, and conservative closed-loop verification contribute?
\textbf{RQ4:} Does ScenePilot improve scenes conservatively, rather than obtaining better metrics through excessive or harmful edits?

\subsection{Experimental Setup}
\label{sec:exp_setup}

We evaluate ScenePilot on SceneEval \citep{tam2026sceneeval} using five heterogeneous text-driven 3D scene generators: ReSpace, LayoutVLM, Reason-3D, LayoutGPT, and InstructScene. None of these generators or their outputs are used to train the repair policy. At evaluation time, the same frozen ScenePilot checkpoint is applied to all five generators without backend-specific policy fine-tuning; only lightweight adapters map each native scene representation and editing interface to the shared canonical interface. Because different generators may produce different object inventories and scene distributions, we evaluate ScenePilot through within-generator paired comparison, i.e., $\mathrm{Base}_g$ versus $\mathrm{Base}_g+\mathrm{ScenePilot}$, rather than comparing absolute scores across generators. We report the official SceneEval Fidelity and Plausibility metrics together with a compact Overall summary score. 
Visual-semantic quality is evaluated separately using a blind GPT-5.4 \citep{openai2026gpt54thinking} judge on layout coherence, relation fidelity, and object completeness, complemented by a blinded human study under the same criteria.

\subsection{Main Results}
\label{sec:main_results}


Table~\ref{tab:main_cross_generator} reports the complete SceneEval
fidelity and plausibility metrics, organized under the two benchmark
categories, together with the Overall summary score. Because different
generators may instantiate different object inventories, we evaluate
ScenePilot through paired improvement over each generator's native
output rather than by directly comparing absolute scores across
generators. Importantly, the repair policy is trained without
generator-specific fine-tuning and is applied to all evaluated
generators through their canonical interface adapters. These results
therefore evaluate whether a single repair policy can transfer across
heterogeneous 3D scene generation pipelines.

\subsection{Ablation Studies and Analysis}

\noindent\textbf{Effect of policy learning.}
Table~\ref{tab:policy_comparison} isolates the effect of policy learning by comparing the pretrained Qwen3-VL-4B backbone, ScenePilot after SFT, the final SFT+GRPO policy, and an expert GPT-5.4 repair agent on the same frozen ReSpace outputs.

\noindent\textbf{Perceptual evaluation.}
Table~\ref{tab:vlm_score} reports 1--10 perceptual scores from two independent VLM judges, GPT-5.4 and Gemini-3-Flash, on \emph{layout coherence}, \emph{relation fidelity}, and \emph{object completeness}. Higher is better. We also conduct a blinded human study under the same criteria, and the rank-based analysis is detailed in Appendix.

\noindent\textbf{Qualitative comparisons.}
Figure~\ref{fig:qualitative_compare} provides a qualitative illustration of this cross-generator repair behavior under the same  instruction. Additional qualitative comparisons across generators are provided in Appendix.

\section{Conclusion}

We present \textbf{ScenePilot}, a plug-and-play framework for generator-agnostic 3D scene repair. By decoupling repair reasoning from generator-specific execution through a canonical interface, learning transferable repair behavior from SceneReverse, and applying conservative closed-loop verification, ScenePilot enables one shared policy to refine scenes from heterogeneous generators without backend-specific policy fine-tuning. Experiments on SceneEval show consistent improvements across five generators unseen during policy training, supporting 3D scene repair as a reusable capability rather than a generator-specific component. This work paves the way for generalizable 3D scene repair across different 3D scene generators.

\noindent\textbf{Limitations and Future Work.}
ScenePilot currently focuses on a compact set of object-level geometric edits, limiting its ability to address errors that require richer structural changes. In addition, each new generator still requires a lightweight adapter for representation conversion and action execution. Future work could explore more expressive canonical operations, more automated adapter construction, and more efficient verification, while extending evaluation to broader scene domains and generation systems.

\section*{LLM Usage Statement}
Generative AI tools were used during the preparation of this work to assist with writing and polishing, retrieval and discovery of relevant literature, coding and debugging. All AI-generated suggestions, code-related recommendations, and retrieved information were critically reviewed and verified by the authors. We take responsibility for the final content of this work.

\bibliographystyle{ieeenat_fullname}
\bibliography{references}

@article{wang2018deeppriors,
  author       = {Kai Wang and
                  Manolis Savva and
                  Angel X. Chang and
                  Daniel Ritchie},
  title        = {Deep convolutional priors for indoor scene synthesis},
  journal      = {{ACM} Trans. Graph.},
  volume       = {37},
  number       = {4},
  pages        = {70},
  year         = {2018},
}

@article{zhao2026scenerevis,
  title   = {{SceneReVis}: A Self-Reflective Vision-Grounded Framework for {3D} Indoor Scene Synthesis via Multi-turn {RL}},
  author  = {Zhao, Yang and Sun, Shizhao and Zhang, Meisheng and Shi, Yingdong and Yang, Xubo and Bian, Jiang},
  journal = {arXiv preprint arXiv:2602.09432},
  year    = {2026},
}

@inproceedings{ritchie2019fastflex,
  title={Fast and Flexible Indoor Scene Synthesis via Deep Convolutional Generative Models},
  author={Ritchie, Daniel and Wang, Kai and Lin, Yu-an},
  booktitle={Proceedings of the IEEE/CVF Conference on Computer Vision and Pattern Recognition},
  year={2019}
}

@inproceedings{wang2021sceneformer,
  title={SceneFormer: Indoor Scene Generation with Transformers},
  author={Wang, Xinpeng and Yeshwanth, Chandan and Nie{\ss}ner, Matthias},
  booktitle={International Conference on 3D Vision},
  year={2021}
}

@inproceedings{zhang2024berfscene,
  title={BerfScene: Bev-conditioned Equivariant Radiance Fields for Infinite 3D Scene Generation},
  author={Zhang, Qihang and Xu, Yinghao and Shen, Yujun and Dai, Bo and Zhou, Bolei and Yang, Ceyuan},
  booktitle={Proceedings of the IEEE/CVF Conference on Computer Vision and Pattern Recognition},
  year={2024}
}

@inproceedings{paschalidou2021atiss,
  title={ATISS: Autoregressive Transformers for Indoor Scene Synthesis},
  author={Paschalidou, Despoina and Kar, Amlan and Shugrina, Maria and Kreis, Karsten and Geiger, Andreas and Fidler, Sanja},
  booktitle={Advances in Neural Information Processing Systems},
  year={2021}
}

@article{zhang2026zonemaestro,
  title={Orchestrating Spatial Semantics via a Zone-Graph Paradigm for Intricate Indoor Scene Generation},
  author={Zhang, Meisheng and Sun, Shizhao and Zhao, Yang and Liu, Ziyuan and Gao, Zhijun and Bian, Jiang},
  journal={arXiv preprint arXiv:2605.02537},
  year={2026}
}

@inproceedings{deng2025global,
  title={Global-Local Tree Search in VLMs for 3D Indoor Scene Generation},
  author={Deng, Wei and Qi, Mengshi and Ma, Huadong},
  booktitle={Proceedings of the IEEE/CVF Conference on Computer Vision and Pattern Recognition},
  year={2025}
}

@inproceedings{tam2026sceneeval,
  title={{SceneEval}: Evaluating Semantic Coherence in Text-Conditioned {3D} Indoor Scene Synthesis},
  author={Tam, Hou In Ivan and Pun, Hou In Derek and Wang, Austin T. and Chang, Angel X. and Savva, Manolis},
  booktitle={Proceedings of the IEEE/CVF Winter Conference on Applications of Computer Vision},
  year={2026}
}

@inproceedings{lin2026pat3d,
  title={PAT3D: Physics-Augmented Text-to-3D Scene Generation},
  author={Lin, Guying and Huang, Kemeng and Liu, Michael and Gao, Ruihan and Chen, Hanke and Chen, Lyuhao and Lu, Beijia and Komura, Taku and Liu, Yuan and Zhu, Jun-Yan and Li, Minchen},
  booktitle={International Conference on Learning Representations},
  year={2026}
}

@inproceedings{chen2025graph2scene,
  title     = {Graph2Scene: Versatile 3D Indoor Scene Generation with Interaction-aware Scene Graph},
  author    = {Chen, Minglin and Yang, Rongkun and Hu, Qibin and Xue, Kaiwen and Zhou, Shunbo and Guo, Yulan},
  booktitle = {IEEE/RSJ International Conference on Intelligent Robots and Systems},
  pages     = {11313--11320},
  year      = {2025},
}

@article{feng2023layoutgpt,
  title={LayoutGPT: Compositional Visual Planning and Generation with Large Language Models},
  author={Feng, Weixi and Zhu, Wanrong and Fu, Tsu-Jui and Jampani, Varun and Akula, Arjun and He, Xuehai and Basu, Sreyashi and Wang, Xin Eric and Ma, William Yang and Krishna, Ranjay and others},
  journal={arXiv preprint arXiv:2305.15393},
  year={2023}
}

@inproceedings{fu2021front,
  title={3D-FRONT: 3D Furnished Rooms with Layouts and Semantics},
  author={Fu, Huan and Cai, Bowen and Gao, Lin and Zhang, Lingxiao and Wang, Cao and Li, Hongbo and Zeng, Yiyun and Sun, Chengyue and Jia, Rongfei and Zhao, Binqiang and others},
  booktitle={Proceedings of the IEEE/CVF International Conference on Computer Vision},
  year={2021}
}

@article{wan2026nala,
  title={NaLA: A 3D Native LLM Layout Agent for High-quality 3D Scene Generation},
  author={Wan, Cheng and Mao, Yongsen and Wu, Wenzheng and Xie, Yuxuan and Xiang, Chucheng and Wang, Runze and Zhang, Xiang and Liu, Zhongyuan and Dai, Rushi and Liu, Yuan},
  journal={arXiv preprint arXiv:2606.29395},
  year={2026}
}

@article{shi2025scenemaker,
  title={SceneMaker: Open-set 3D Scene Generation with Decoupled De-occlusion and Pose Estimation Model},
  author={Shi, Yukai and Li, Weiyu and Wang, Zihao and Li, Hongyang and Chen, Xingyu and Tan, Ping and Zhang, Lei},
  journal={arXiv preprint arXiv:2512.10957},
  year={2025}
}

@article{zhou2025holodreamer,
  title={HoloDreamer: Holistic 3D Panoramic Scene Generation From Text Descriptions},
  author={Zhou, Haiyang and Cheng, Xinhua and Yu, Wangbo and Tian, Yonghong and Yuan, Li},
  journal={IEEE Transactions on Visualization and Computer Graphics},
  year={2025},
}

@inproceedings{wang2019planit,
  title     = {{PlanIT}: Planning and Instantiating Indoor Scenes with Relation Graph and Spatial Prior Networks},
  author    = {Wang, Kai and Lin, Yu-An and Weissmann, Benjamin and Savva, Manolis and Chang, Angel X. and Ritchie, Daniel},
  booktitle = {ACM Transactions on Graphics},
  volume    = {38},
  number    = {4},
  pages     = {1--15},
  year      = {2019}
}

@inproceedings{deitke2022procthor,
  title     = {{ProcTHOR}: Large-Scale Embodied AI Using Procedural Generation},
  author    = {Deitke, Matt and VanderBilt, Eli and Herrasti, Alvaro and Weihs, Luca and Salvador, Jordi and Ehsani, Kiana and Kolve, Eric and Farhadi, Ali and Kembhavi, Aniruddha and Mottaghi, Roozbeh},
  booktitle = {Advances in Neural Information Processing Systems},
  volume    = {35},
  pages     = {5982--5994},
  year      = {2022}
}

@inproceedings{yang2024holodeck,
  title={Holodeck: Language Guided Generation of 3D Embodied AI Environments},
  author={Yang, Yue and Sun, Fan-Yun and Weihs, Luca and Vanderbilt, Eli and Herrasti, Alvaro and Han, Winson and Wu, Jiajun and Haber, Nick and Krishna, Ranjay and Liu, Lingjie and others},
  booktitle={Proceedings of the IEEE/CVF Conference on Computer Vision and Pattern Recognition},
  year={2024}
}

@inproceedings{yang2024physcene,
  title={PhyScene: Physically Interactable 3D Scene Synthesis for Embodied AI},
  author={Yang, Yandan and Jia, Baoxiong and Zhi, Peiyuan and Huang, Siyuan},
  booktitle={Proceedings of the IEEE/CVF Conference on Computer Vision and Pattern Recognition},
  year={2024}
}

@inproceedings{tang2024diffuscene,
  title={DiffuScene: Denoising Diffusion Models for Generative Indoor Scene Synthesis},
  author={Tang, Jiapeng and Nie, Yinyu and Markhasin, Lev and Dai, Angela and Thies, Justus and Nie{\ss}ner, Matthias},
  booktitle={Proceedings of the IEEE/CVF Conference on Computer Vision and Pattern Recognition},
  year={2024}
}

@inproceedings{maillard2024debara,
  title={DeBaRA: Denoising-Based 3D Room Arrangement Generation},
  author={Maillard, L{\'e}opold and Sereyjol-Garros, Nicolas and Durand, Tom and Ovsjanikov, Maks},
  booktitle={Advances in Neural Information Processing Systems},
  year={2024}
}

@article{bucher2025respace,
  title={Respace: Text-driven 3d scene synthesis and editing with preference alignment},
  author={Bucher, Martin JJ and Armeni, Iro},
  journal={arXiv e-prints},
  pages={arXiv--2506},
  year={2025}
}

@article{pan2025metaspatial,
  title={MetaSpatial: Reinforcing 3D Spatial Reasoning in VLMs for the Metaverse},
  author={Pan, Zhenyu and Liu, Han},
  journal={arXiv preprint arXiv:2503.18470},
  year={2025}
}

@article{sun2025layoutvlm,
  title={LayoutVLM: Differentiable Optimization of 3D Layout via Vision-Language Models},
  author={Sun, Fan-Yun and Liu, Weiyu and Gu, Siyi and Lim, Dylan and Bhat, Goutam and Tombari, Federico and Li, Manling and Haber, Nick and Wu, Jiajun},
  journal={arXiv preprint arXiv:2412.02193},
  year={2025}
}

@inproceedings{yang2025mmgdreamer,
  title={MMGDreamer: Mixed-Modality Graph for Geometry-Controllable 3D Indoor Scene Generation},
  author={Yang, Zhifei and Lu, Keyang and Zhang, Chao and Qi, Jiaxing and Jiang, Hanqi and Ma, Ruifei and Yin, Shenglin and Xu, Yifan and Xing, Mingzhe and Xiao, Zhen and Long, Jieyi and Liu, Xiangde and Zhai, Guangyao},
  booktitle={Proceedings of the AAAI Conference on Artificial Intelligence},
  volume={39},
  number={9},
  pages={9391--9399},
  year={2025}
}

@inproceedings{yi2022mime,
  title={{MIME}: Human-Aware {3D} Scene Generation},
  author={Yi, Hongwei and Huang, Chun-Hao P. and Tripathi, Shashank and Hering, Lea and Thies, Justus and Black, Michael J.},
  booktitle={Proceedings of the IEEE/CVF Conference on Computer Vision and Pattern Recognition},
  pages={12965--12976},
  year={2023}
}

@inproceedings{gu2025artiscene,
  title={ArtiScene: Language-Driven Artistic 3D Scene Generation Through Image Intermediary},
  author={Gu, Zeqi and Cui, Yin and Li, Zhaoshuo and Wei, Fangyin and Ge, Yunhao and Gu, Jinwei and Liu, Ming-Yu and Davis, Abe and Ding, Yifan},
  booktitle={Proceedings of the IEEE/CVF Conference on Computer Vision and Pattern Recognition},
  year={2025}
}

@article{ran2025directlayout,
  title={Direct Numerical Layout Generation for 3D Indoor Scene Synthesis via Spatial Reasoning},
  author={Ran, Xingjian and Li, Yixuan and Xu, Linning and Yu, Mulin and Dai, Bo},
  journal={arXiv preprint arXiv:2506.05341},
  year={2025}
}

@article{yang2025sceneweaver,
  title={SceneWeaver: All-in-One 3D Scene Synthesis with an Extensible and Self-Reflective Agent},
  author={Yang, Yandan and Jia, Baoxiong and Zhang, Shujie and Huang, Siyuan},
  journal={arXiv preprint arXiv:2509.20414},
  year={2025}
}

@article{song2023roomdreamer,
  title={RoomDreamer: Text-Driven 3D Indoor Scene Synthesis with Coherent Geometry and Texture},
  author={Song, Liangchen and Cao, Liangliang and Xu, Hongyu and Kang, Kai and Tang, Feng and Yuan, Junsong and Zhao, Yang},
  journal={arXiv preprint arXiv:2305.11337},
  year={2023}
}

@inproceedings{bautista2022gaudi,
  title={GAUDI: A Neural Architect for Immersive 3D Scene Generation},
  author={Bautista, Miguel Angel and Guo, Pengsheng and Abnar, Samira and Talbott, Walter and Toshev, Alexander and Chen, Zhuoyuan and Dinh, Laurent and Zhai, Shuangfei and Goh, Hanlin and Ulbricht, Daniel and Dehghan, Afshin and Susskind, Joshua},
  booktitle={Advances in Neural Information Processing Systems},
  volume={35},
  pages={25102--25116},
  year={2022}
}

@inproceedings{zhou2025scenex,
  title     = {SceneX: Procedural Controllable Large-Scale Scene Generation},
  author    = {Zhou, Mengqi and Wang, Yuxi and Hou, Jun and Zhang, Shougao and Li, Yiwei and Luo, Chuanchen and Peng, Junran and Zhang, Zhaoxiang},
  booktitle = {Proceedings of the AAAI Conference on Artificial Intelligence},
  volume    = {39},
  number    = {10},
  pages     = {10806--10814},
  year      = {2025},
  doi       = {10.1609/aaai.v39i10.33174}
}

@inproceedings{ocal2024sceneteller,
  title     = {SceneTeller: Language-to-3D Scene Generation},
  author    = {{\"O}cal, Ba{\c{s}}ak Melis and Tatarchenko, Maxim and Karao{\u{g}}lu, Sezer and Gevers, Theo},
  booktitle = {European Conference on Computer Vision},
  pages     = {362--378},
  year      = {2024},
  publisher = {Springer}
}

@inproceedings{celen2025idesign,
  title={I-design: Personalized llm interior designer},
  author={{\c{C}}elen, Ata and Han, Guo and Schindler, Konrad and Van Gool, Luc and Armeni, Iro and Obukhov, Anton and Wang, Xi},
  booktitle={European Conference on Computer Vision},
  pages={217--234},
  year={2024},
  organization={Springer}
}

@article{berdoz2025reason3d,
  title={Text-to-Scene with Large Reasoning Models},
  author={Berdoz, Fr{\'e}d{\'e}ric and Lanzend{\"o}rfer, Luca A. and Tuninga, Nick and Wattenhofer, Roger},
  journal={arXiv preprint arXiv:2509.26091},
  year={2025}
}

@article{shao2024grpo,
  title={DeepSeekMath: Pushing the Limits of Mathematical Reasoning in Open Language Models},
  author={Shao, Zhihong and Wang, Peiyi and Zhu, Qihao and Xu, Runxin and Song, Junxiao and Bi, Xiao and Zhang, Haowei and Zhang, Mingchuan and Li, Y. K. and Wu, Y. and others},
  journal={arXiv preprint arXiv:2402.03300},
  year={2024}
}

@inproceedings{yao2023react,
  title     = {ReAct: Synergizing Reasoning and Acting in Language Models},
  author    = {Yao, Shunyu and Zhao, Jeffrey and Yu, Dian and Du, Nan and Shafran, Izhak and Narasimhan, Karthik and Cao, Yuan},
  booktitle = {International Conference on Learning Representations},
  year      = {2023},
}

@misc{openai2026gpt54thinking,
  title        = {{GPT-5.4 Thinking System Card}},
  author       = {{OpenAI}},
  year         = {2026},
  month        = mar,
  howpublished = {\url{https://openai.com/index/gpt-5-4-thinking-system-card/}}
}

@article{lin2024instructscene,
  title={InstructScene: Instruction-Driven 3D Indoor Scene Synthesis with Semantic Graph Prior},
  author={Lin, Chenguo and Mu, Yadong},
  journal={arXiv preprint arXiv:2402.04717},
  year={2024}
}

@inproceedings{fang2025ctrlroom,
  title={Ctrl-Room: Controllable Text-to-3D Room Meshes Generation with Layout Constraints},
  author={Fang, Chuan and Dong, Yuan and Luo, Kunming and Hu, Xiaotao and Shrestha, Rakesh and Tan, Ping},
  booktitle={International Conference on 3D Vision},
  pages={692--701},
  year={2025},
  publisher={IEEE}
}

@inproceedings{madaan2023selfrefine,
  title     = {Self-Refine: Iterative Refinement with Self-Feedback},
  author    = {Madaan, Aman and Tandon, Niket and Gupta, Prakhar and Hallinan, Skyler and Gao, Luyu and Wiegreffe, Sarah and Alon, Uri and Dziri, Nouha and Prabhumoye, Shrimai and Yang, Yiming and Gupta, Shashank and Majumder, Bodhisattwa Prasad and Hermann, Katherine and Welleck, Sean and Yazdanbakhsh, Amir and Clark, Peter},
  booktitle = {Advances in Neural Information Processing Systems},
  year      = {2023},
}

@article{xia2026sage,
  title   = {SAGE: Scalable Agentic 3D Scene Generation for Embodied AI},
  author  = {Xia, Hongchi and Li, Xuan and Li, Zhaoshuo and Ma, Qianli and Xu, Jiashu and Liu, Ming-Yu and Cui, Yin and Lin, Tsung-Yi and Ma, Wei-Chiu and Wang, Shenlong and Song, Shuran and Wei, Fangyin},
  journal = {arXiv preprint arXiv:2602.10116},
  year    = {2026},
}

@article{ling2025scenethesis,
  title   = {Scenethesis: A Language and Vision Agentic Framework for 3D Scene Generation},
  author  = {Ling, Lu and Lin, Chen-Hsuan and Lin, Tsung-Yi and Ding, Yifan and Zeng, Yu and Sheng, Yichen and Ge, Yunhao and Liu, Ming-Yu and Bera, Aniket and Li, Zhaoshuo},
  journal = {arXiv preprint arXiv:2505.02836},
  year    = {2025}
}

@article{qwen2025qwen3vl,
  title   = {{Qwen3-VL} Technical Report},
  author  = {Bai, Shuai and others},
  journal = {arXiv preprint arXiv:2511.21631},
  year    = {2025},
}

\clearpage

\appendix

\setcounter{figure}{0}
\setcounter{table}{0}
\setcounter{equation}{0}
\setcounter{algorithm}{0}

\renewcommand{\thefigure}{S\arabic{figure}}
\renewcommand{\thetable}{S\arabic{table}}
\renewcommand{\theequation}{S\arabic{equation}}
\renewcommand{\thealgorithm}{S\arabic{algorithm}}

\renewcommand{\theHfigure}{supp.\arabic{figure}}
\renewcommand{\theHtable}{supp.\arabic{table}}
\renewcommand{\theHequation}{supp.\arabic{equation}}

\providecommand{\theHalgorithm}{}
\renewcommand{\theHalgorithm}{supp.\arabic{algorithm}}

\maketitlesupplementary

\let\addcontentsline\standardaddcontentsline
\tableofcontents

\section{Experimental Details}
\label{app:experimental_details}

\subsection{Model and Training Setup}
\label{app:training_setup}

\paragraph{Base model and observation format.}
We instantiate the shared repair policy with Qwen3-VL-4B-Instruct. Each policy input contains two rendered observations of the current scene---a diagonal view and an annotated top-down view---together with the canonical structured scene state and the capability descriptor exposed by the current backend. The model predicts exactly one atomic canonical decision from
\[
\begin{aligned}
\mathcal{A}\cup\{\texttt{STOP}\},\qquad
\mathcal{A}
&= \{\texttt{translate},\texttt{rotate},\\
&\quad \texttt{scale},\texttt{delete},\texttt{add}\}.
\end{aligned}
\]
The output is constrained to a strict JSON schema and contains no free-form explanation. Generator identity is not provided as an explicit policy input.

\paragraph{Supervised fine-tuning.}
The final SFT stage uses the room-disjoint SceneReverse SFT partition described in Appendix~\ref{app:dataset_statistics}. We train for one epoch with LoRA rank $64$, LoRA scaling factor $128$, and dropout $0.05$. Optimization uses AdamW with learning rate $2\times10^{-5}$, weight decay $0.1$, cosine learning-rate decay, warmup ratio $0.03$, and bf16 precision. The per-device batch size is $1$, and gradient accumulation is chosen to maintain an effective global batch size of approximately $32$. We use a maximum sequence length of $5632$ tokens and FlashAttention-2. Only assistant output tokens are supervised; image tokens and input prompt tokens are masked from the language-model loss. 

\paragraph{GRPO initialization.}
The GRPO stage is initialized from the final SFT checkpoint and uses only the disjoint GRPO partition of SceneReverse. Each sampled candidate is parsed through the same canonical schema and capability-aware execution interface used at inference. The final GRPO reward is
\begin{equation}
\begin{aligned}
R
&= \lambda_{\mathrm{fmt}}R_{\mathrm{format}}
 + \lambda_{\mathrm{phy}}\Delta Q_{\mathrm{PHY}} \\
&\quad + \lambda_{\mathrm{spa}}\Delta Q_{\mathrm{SPA}}
 + \lambda_{\mathrm{vlm}}\Delta Q_{\mathrm{VLM}}.
\end{aligned}
\label{eq:app_grpo_reward}
\end{equation}
where $R_{\mathrm{format}}$ jointly measures schema validity and executability, while the remaining terms measure the realized scene-quality change after applying the candidate edit. We do not use a ground-truth action-matching reward. The final rollout and optimizer hyperparameters are reported from the frozen training run in Table~\ref{tab:grpo_hparams}.

\begin{table}[h]
\centering
\caption{Hyperparameters used in the final GRPO training run.}
\label{tab:grpo_hparams}
\small
\begin{tabular}{lr}
\toprule
Setting & Value \\
\midrule
GRPO states & 8,574 \\
Rollouts per state & 4 \\
Learning rate & $5\times10^{-6}$ \\
Epochs & 1 \\
$\lambda_{\mathrm{fmt}}$ & 0.30 \\
$\lambda_{\mathrm{phy}}$ & 0.40 \\
$\lambda_{\mathrm{spa}}$ & 0.15 \\
$\lambda_{\mathrm{vlm}}$ & 0.15 \\
KL coefficient & 0.04 \\
\bottomrule
\end{tabular}
\end{table}

\paragraph{Implementation environment.}
Training is conducted on a Linux server with two NVIDIA RTX A6000 GPUs (48\,GB each), while evaluation is performed on a separate Linux server with four NVIDIA RTX 4090D GPUs (24\,GB each). Rendering uses the backend-native scene assets and the same renderer for the unrepaired and repaired versions of each generator output. Unless otherwise noted, the reported ScenePilot refinement time excludes the runtime of the original scene generator.

\subsection{SceneReverse Data Statistics}
\label{app:dataset_statistics}

The final SceneReverse pool contains $9{,}582$ unique rooms, $20{,}012$ validated recovery trajectories, and $53{,}599$ atomic recovery states. We split the pool by room identity, so no room appears in both SFT and GRPO. Table~\ref{tab:dataset_stats} summarizes the final split.

\begin{table}[h]
\centering
\caption{Statistics of the final room-disjoint SceneReverse training split.}
\label{tab:dataset_stats}
\small
\begin{tabular}{lrrr}
\toprule
Item & Full pool & SFT & GRPO \\
\midrule
Rooms & 9,582 & 8,050 & 1,532 \\
Trajectories & 20,012 & 16,810 & 3,202 \\
Recovery steps & 53,599 & 45,025 & 8,574 \\
Terminal \texttt{STOP} samples & N/A & 4,246 & N/A \\
Total SFT samples & N/A & 49,271 & N/A \\
\bottomrule
\end{tabular}
\end{table}

Table~\ref{tab:action_distribution} reports the decision distribution in the final SFT set. Each sample contains exactly one atomic target decision.

\begin{table}[h]
\centering
\caption{Atomic decision distribution in the final SceneReverse SFT set.}
\label{tab:action_distribution}
\small
\begin{tabular}{lr}
\toprule
Decision & Samples \\
\midrule
\texttt{translate} & 15,161 \\
\texttt{rotate} & 13,783 \\
\texttt{scale} & 2,297 \\
\texttt{delete} & 6,892 \\
\texttt{add} & 6,892 \\
\texttt{STOP} & 4,246 \\
\midrule
Total & 49,271 \\
\bottomrule
\end{tabular}
\end{table}

\subsection{Capability-Conditioned SFT Sampling}
\label{app:capability_sampling}

To prevent the SFT policy from assuming that every backend supports the full canonical action set, we augment each SFT sample with a capability descriptor. The final capability sampler produces approximately equal numbers of full-capability and partial-capability examples. Of the $49{,}271$ SFT samples, $24{,}806$ expose all five edit operations and $24{,}465$ expose a four-operation subset. For a non-terminal target, partial-capability sampling disables exactly one non-target operation, ensuring that the demonstrated target action remains executable. For a terminal \texttt{STOP} sample, one edit operation is disabled while \texttt{STOP} remains available. This construction exposes the policy to heterogeneous action interfaces without revealing generator identity.

\subsection{Benchmark and Evaluation Protocol}
\label{app:benchmark_protocol}

\paragraph{SceneEval protocol.}
We evaluate ScenePilot on a fixed set of $100$ SceneEval prompts across five scene generators: ReSpace, LayoutVLM, Reason-3D, LayoutGPT, and InstructScene. All generators receive the same prompts and logical room specifications while retaining their native object-selection mechanisms, asset inventories, and scene representations. The shared repair policy is trained on SceneReverse rather than generator-specific failure trajectories. The same frozen policy checkpoint is applied to all five generators without generator-specific fine-tuning. None of the five evaluated generators is used for policy training or hyperparameter tuning.

\paragraph{Paired comparison.}
Because heterogeneous generators may instantiate different object inventories for the same prompt, our primary comparison is paired within each backend:
\[
\mathrm{Base}_g
\quad\text{vs.}\quad
\mathrm{Base}_g+\mathrm{ScenePilot}.
\]
We therefore emphasize generator-level improvements and macro-average these improvements across backends rather than treating absolute scores from incompatible object sets as directly interchangeable.

\paragraph{SceneEval metrics.}
We use the official SceneEval evaluator.
The paper-level Fidelity, Plausibility, and Overall scores are computed exactly by Eqs.~(\ref{eq:sceneeval_fidelity_aggregate}--\ref{eq:sceneeval_overall}).
The Fidelity aggregate uses CNT, ATR, OOR, and OAR.
The Plausibility aggregate uses object-level collision, support, navigability, accessibility, and out-of-boundary rate, with lower-is-better terms complemented before averaging.
We additionally retain $\mathrm{COL}_{\mathrm{sc}}$ in the detailed results below but exclude it from the aggregate to avoid assigning collision two weights.

\paragraph{VLM-based evaluation.}
For final visual-semantic evaluation, we use GPT-5.4
at temperature $0$ in a blind single-scene protocol.
The judge receives the rendered scene observations
and the corresponding task specification without
access to the generator identity, repair policy,
or paired counterpart. The judge assigns integer
scores from $1$ to $10$ for layout coherence,
relation fidelity, and object completeness.
Missing scores are retained as missing values
rather than replaced by heuristic estimates.

\section{Additional Quantitative Results}
\label{app:additional_quantitative}

\subsection{Detailed VLM-Judge Results}
\label{app:vlm_results}

\begin{table}[h]
\centering
\caption{Detailed blind VLM-judge scores before conversion to ranks. Scores are integers in $[1,10]$ at the scene level; the table reports means across valid scenes.}
\label{tab:vlm_dimensions}
\footnotesize
\setlength{\tabcolsep}{3pt}
\begin{tabular}{lcccc}
\toprule
Policy
& Layout$\uparrow$
& Relation$\uparrow$
& Completeness$\uparrow$
& Avg.$\uparrow$ \\
\midrule
Qwen3-VL-4B  & 6.92 & 6.59 & 6.83 & 6.78 \\
ScenePilot-SFT & 7.35 & 6.65 & 6.85 & 6.95 \\
ScenePilot-SFT+GRPO & 7.57 & 6.93 & 7.01 & 7.17 \\
w/o ScenePilot & 7.21 & 6.81 & 6.92 & 6.98 \\
Expert-GPT-5.4$^\dagger$ & 7.24 & 7.19 & 7.57 & 7.33 \\
\bottomrule
\end{tabular}
\end{table}

\subsection{Human--VLM Agreement}
\label{app:human_vlm_agreement}
We use a blinded human study as an auxiliary validation of the VLM-based evaluator. We recruit 15 graduate students to rank anonymized policy outputs under the same three criteria used by the VLM judges: \emph{layout coherence}, \emph{relation fidelity}, and \emph{object completeness}. We report average human ranks and compute Pearson's $r$
and Spearman's $\rho$ between the human and VLM mean
ranks across the four evaluated conditions for each
criterion. These correlations characterize agreement at
the condition level rather than the individual-scene level.

\begin{table}[t]
\centering
\caption{\textbf{Average ranks based on blinded user ratings and VLM scores.}
Lower rank indicates better perceptual quality. Each entry reports the User rank followed by the VLM rank.}
\label{tab:user_vlm_rank}
\footnotesize
\setlength{\tabcolsep}{3pt}
\begin{tabular}{lccc}
\toprule
Policy
& \makecell{Layout\\(User/VLM)}
& \makecell{Relation\\(User/VLM)}
& \makecell{Completeness\\(User/VLM)} \\
\midrule
ReSpace
& 2.31 / 2.21 & 2.19 / 2.48 & 2.07 / 2.18 \\

w/ ScenePilot (ours)
& 1.62 / 1.73 & 1.52 / 1.68 & 1.73 / 1.68 \\

LayoutVLM
& 3.34 / 3.08 & 3.27 / 3.38 & 3.08 / 3.10 \\

w/ ScenePilot (ours)
& 2.72 / 2.98 & 3.02 / 2.46 & 3.11 / 3.05 \\
\bottomrule
\end{tabular}
\end{table}

\begin{table}[h]
\centering
\caption{\textbf{Agreement between blinded user ratings and VLM scores.}}
\label{tab:user_vlm_correlation}
\small
\setlength{\tabcolsep}{10pt}
\begin{tabular}{lcc}
\toprule
Criterion & Pearson $r$ & Spearman $\rho$ \\
\midrule
Layout coherence    & 0.950 & 1.000 \\
Relation fidelity   & 0.879 & 0.800 \\
Object completeness & 0.994 & 0.800 \\
\bottomrule
\end{tabular}
\end{table}

\section{Additional Details of SceneReverse}
\label{app:scenereverse_details}

\subsection{Canonical Scene Representation}
\label{app:scene_representation}

SceneReverse and the repair policy use the same generator-independent, object-centric representation. A canonical scene is written as
\begin{equation}
\begin{aligned}
z &= (B,\mathcal{O}), \qquad
\mathcal{O} = \{o_i\}_{i=1}^{N},\\
o_i &= (\mathrm{id}_i,c_i,\mathbf{p}_i,
\theta_i,\mathbf{s}_i).
\end{aligned}
\label{eq:canonical_scene_representation}
\end{equation}
contains a stable object identifier, semantic category, metric position, yaw angle, and physical size. Backend adapters normalize native coordinate frames, units, yaw conventions, and room-boundary formats before the state is exposed to the policy.

\IfFileExists{fig/scene_representation_one.pdf}{
\begin{figure}[h]
    \centering
    \includegraphics[width=0.95\linewidth]{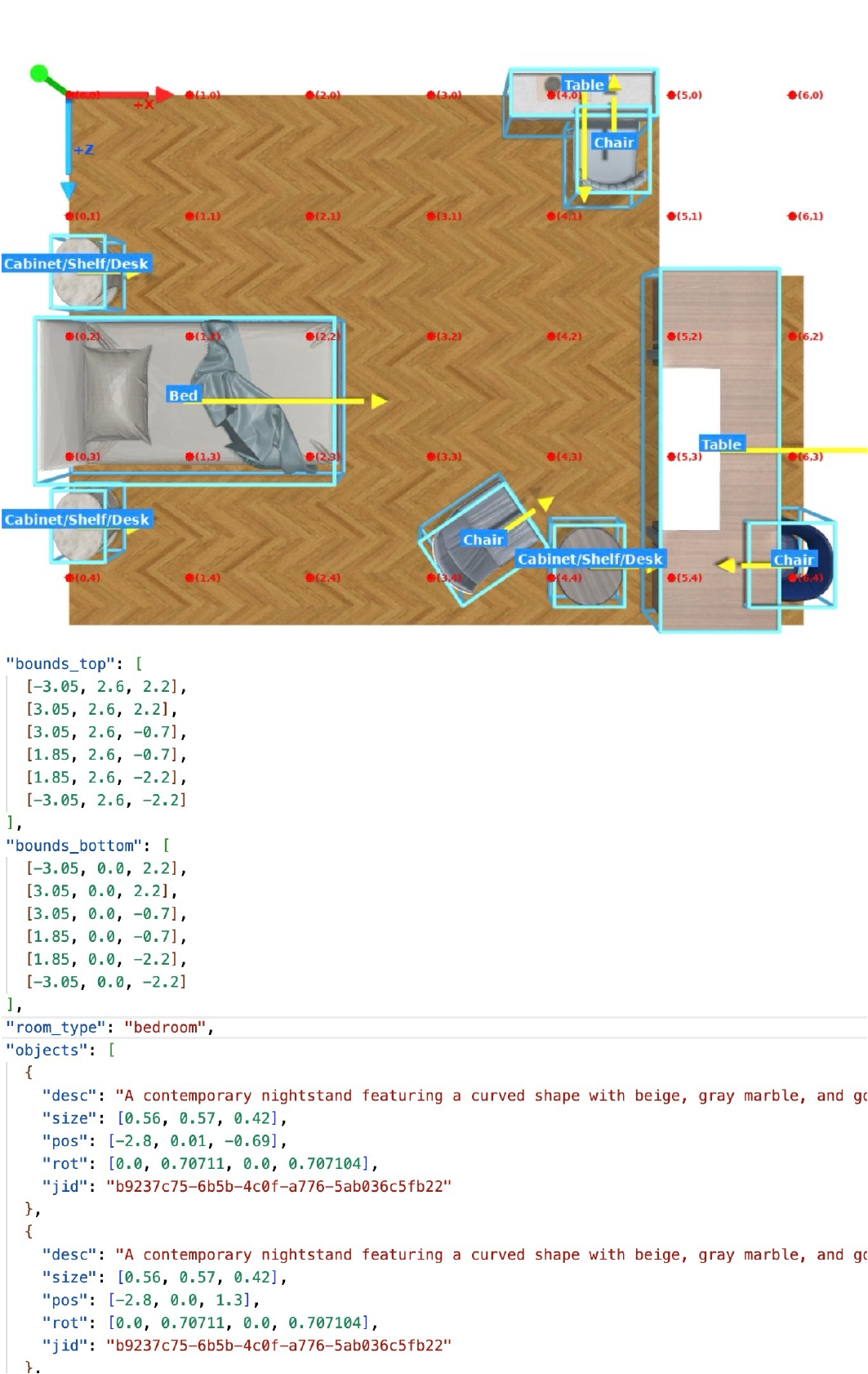}
    \caption{Example of the canonical object-centric scene representation used by ScenePilot.}
    \label{fig:scene_representation}
\end{figure}
}{}

\subsection{SceneReverse Construction and Filtering Details}
\label{app:scenereverse_filtering}

SceneReverse starts from a clean scene $S^\star$ and applies a sequence of structured perturbations in the canonical edit space. Each perturbation has a known inverse, allowing the construction of an executable recovery path. The resulting inverse action should be interpreted as one verifiable recovery action induced by the corruption process, rather than as the unique optimal repair for the degraded scene.

Given a clean scene $S^\star$, we generate degraded intermediate states through executable perturbations:
\begin{equation}
    \hat{S}_0=S^\star,\qquad
    \hat{S}_t=c_t(\hat{S}_{t-1};\beta_t),\qquad
    c_t\sim\mathcal{C},
    \label{eq:structured_corruption}
\end{equation}
where $\mathcal{C}$ denotes the set of reversible perturbations defined in the canonical action space, and $\beta_t$ controls the perturbation magnitude.

Starting from the most degraded state $S_0^{\mathrm{r}}=\hat{S}_T$, the recovery trajectory is obtained by reversing the applied perturbations:
\begin{equation}
\begin{aligned}
a_t^\star
&= \operatorname{Inv}(c_{T-t}),\\
S_{t+1}^{r}
&= f_{\mathrm{apply}}(S_t^{r},a_t^\star),
\quad t=0,\ldots,T-1.
\end{aligned}
\label{eq:reverse_recovery}
\end{equation}
To remove nearly unchanged transitions, we measure the normalized magnitude of the state change. For matched objects, the per-step geometric change can be written as
\begin{equation}
\begin{aligned}
V_t
&= \frac{1}{|\mathcal{O}_t|}
\sum_{o_i\in\mathcal{O}_t}
\Bigg[
 w_p\frac{\|\Delta\mathbf{p}_{i,t}\|_2}{d_B+\epsilon}\\
&\qquad + w_r\frac{|\Delta\theta_{i,t}|}{180^\circ}
 + w_s\left\|
 \log\frac{\mathbf{s}_{i,t+1}}{\mathbf{s}_{i,t}}
 \right\|_1
\Bigg].
\end{aligned}
\label{eq:trajectory_volatility}
\end{equation}
where $d_B$ is a room-scale normalization factor derived from the floor-plan extent. This normalization prevents translation, rotation, and scale changes from being compared in incompatible units.

We additionally characterize how far an intermediate state remains from the clean reference. For each matched object, we normalize the remaining error by the magnitude of the original corruption:
\begin{equation}
\begin{aligned}
D_t
&= \frac{1}{|\mathcal{O}_t|}
\sum_{o_i\in\mathcal{O}_t} \\
&\quad \left[
\begin{aligned}
&\alpha_p
\frac{\|\mathbf{p}^{\star}_i-\mathbf{p}_{i,t}\|_2}
{\|\mathbf{p}^{\star}_i-\mathbf{p}^{0}_i\|_2+\epsilon} \\
&+\alpha_r
\frac{|\theta^{\star}_i-\theta_{i,t}|}
{|\theta^{\star}_i-\theta^{0}_i|+\epsilon} \\
&+\alpha_s
\frac{\|\mathbf{s}^{\star}_i-\mathbf{s}_{i,t}\|_2}
{\|\mathbf{s}^{\star}_i-\mathbf{s}^{0}_i\|_2+\epsilon}
\end{aligned}
\right].
\end{aligned}
\label{eq:trajectory_difficulty}
\end{equation}
where $(\mathbf{p}^{0}_i,\theta^{0}_i,\mathbf{s}^{0}_i)$ denotes the initial degraded state and $(\mathbf{p}^{\star}_i,\theta^{\star}_i,\mathbf{s}^{\star}_i)$ denotes the clean target. We use this residual measure only to organize recovery difficulty and data coverage; it is not used as a ground-truth action-matching reward during GRPO.

\subsection{Terminal Supervision}
\label{app:terminal_supervision}

Recovery trajectories provide executable supervision until the clean terminal state is reached. We pair a subset of recovered terminal states with \texttt{STOP}, yielding $4{,}246$ terminal SFT examples.  The terminal examples teach the policy to preserve a satisfactory state rather than applying unnecessary edits.

\section{Generator Adapters and Canonical Interface}
\label{app:adapters}

\subsection{Backend Capability Matrix}

Each generator adapter exposes only the canonical operations that can be grounded into executable native edits. Table~\ref{tab:backend_capabilities} is filled from the final frozen adapters rather than inferred from generator names.

\begin{table}[h]
\centering
\caption{Distribution of atomic ScenePilot decisions during one repair run over the 100 SceneEval cases for each generator. Values denote the number of executed actions; ``--'' indicates that the corresponding operation is not exposed by the backend adapter.}
\label{tab:backend_capabilities}
\footnotesize
\setlength{\tabcolsep}{3pt}
\begin{tabular}{lcccccc}
\toprule
Generator & Translate & Rotate & Scale & Delete & Add & Stop\\
\midrule
ReSpace       & 44 & 35 & 14 & 2 & 9  & 3  \\
LayoutVLM     & 48 & 40 & -- & -- & -- & 12 \\
Reason-3D     & 41 & 30 & -- & 4 & 8  & 7  \\
LayoutGPT     & 23 & 15 & 8  & 8 & 29 & 6  \\
InstructScene & 34 & 31 & 13 & 8 & 11 & 2  \\
\bottomrule
\end{tabular}
\end{table}

\subsection{Adapter Responsibilities}
For backend $g$, the interface $I_g=(\phi_g,\psi_g,C_g)$ has three responsibilities. First, $\phi_g$ canonicalizes the native room geometry, object identifiers, poses, physical sizes, and semantic categories. Second, $\psi_g$ translates one validated canonical action into the backend's native editing representation. Third, $C_g$ exposes the supported operation subset and valid parameter ranges. No generator-specific policy checkpoint is introduced: the same repair-policy parameters are reused across all backends.

The adapter is intentionally restricted to representation conversion and action execution. It does not contain generator-specific repair heuristics, hidden prompts that decide which object should be changed, or backend-specific policy fine-tuning. This separation is essential for interpreting held-out-backend performance as transfer of the shared repair policy rather than as a collection of separately engineered optimizers.

\section{Training and Inference Details}
\label{app:training_inference_details}

\subsection{Step-Level GRPO Credit Assignment}

For each GRPO state $S_t$, the policy samples a group of candidate atomic decisions. A candidate is first parsed and validated. If it is executable and non-terminal, it is applied to a temporary copy of the scene, producing $\widetilde{S}_{t+1}^{(k)}$. Reward components are computed from realized scene deltas
\begin{equation}
\begin{aligned}
\Delta Q_m^{(k)}
&= Q_m\!\left(\widetilde{S}_{t+1}^{(k)}\right)
   - Q_m(S_t),\\
&\qquad m\in\{\mathrm{PHY},\mathrm{SPA},\mathrm{VLM}\}.
\end{aligned}
\label{eq:step_level_reward_delta}
\end{equation}
The group-relative advantage is obtained by normalizing rewards within the sampled group, and the policy is optimized with the standard clipped GRPO objective and KL regularization to the SFT reference policy. Because credit is assigned to individual executable edits, the RL stage does not require an oracle final layout or a unique ground-truth repair action.

\subsection{Verifier Configuration}
\label{app:verifier_configuration}

The verifier evaluates scene quality using the weighted
score defined in the main paper. All three quality
components are normalized to $[0,1]$.

\paragraph{Physical quality.}
We measure physical validity using the proportions of
objects involved in collisions and objects outside
the room boundary:
\begin{equation}
Q_{\mathrm{PHY}}(S)
=
1-\frac{1}{2}
\left(
\frac{N_{\mathrm{collision}}(S)}{N(S)}
+
\frac{N_{\mathrm{oob}}(S)}{N(S)}
\right),
\label{eq:verifier_physical}
\end{equation}
where $N(S)$ is the total number of objects,
$N_{\mathrm{collision}}(S)$ counts objects involved
in at least one collision, and $N_{\mathrm{oob}}(S)$
counts objects outside the room boundary.
Both violation counts are obtained using SceneEval.
A collision-free scene with no boundary violations
receives a score of $1$. Empty scenes are treated
as invalid candidates.

\paragraph{Spatial quality.}
We extract a fixed set of spatial relations
$\mathcal{R}$ from the input instruction and compute
\begin{equation}
Q_{\mathrm{SPA}}(S)
=
\frac{1}{|\mathcal{R}|}
\sum_{r\in\mathcal{R}}s_r(S),
\label{eq:verifier_spatial}
\end{equation}
where $s_r(S)\in[0,1]$ measures the satisfaction
of relation $r$. The supported relation types are
\texttt{near}, \texttt{next\_to}, \texttt{face\_to},
\texttt{side\_of}, \texttt{against\_wall}, and
\texttt{middle\_of\_room}.

Distance-based relations are evaluated using
\begin{equation}
b(d;l,u,\sigma)
=
\exp\left(
-\frac{\operatorname{dist}(d,[l,u])^2}{2\sigma^2}
\right),
\label{eq:relation_distance}
\end{equation}
where $d$ is the measured distance and $[l,u]$
is the desired distance interval.

The parameters $(l,u,\sigma)$ are $(0.5,1.5,0.25)$
for \texttt{near}, $(0,0.5,0.25)$ for
\texttt{next\_to}, and $(0,0.3,0.1)$ for
\texttt{against\_wall}, with distances measured
in meters.

For \texttt{face\_to}, the relation score is
$\max(0,1-\theta/30^\circ)$, where $\theta$
is the angular deviation from the desired orientation.
For \texttt{side\_of}, the score is the fraction
of target oriented bounding-box corners satisfying
the specified directional constraint.
For \texttt{middle\_of\_room}, a Gaussian score
is computed from the distance to the room center,
with its scale determined by room and object sizes.

The relation set and object-instance bindings remain
fixed throughout repair. Relations involving missing
required objects receive a score of $0$.
When no spatial relations are available for evaluation,
we set $Q_{\mathrm{SPA}}(S)=1$.

\paragraph{Visual-semantic quality.}
At inference, GPT-5.4 evaluates the scene using
top-down and diagonal renderings from fixed cameras,
together with the original user instruction.
The judge scores layout correctness, semantic
plausibility, and functional completeness on
a scale of $[0,10]$. We normalize their average as
\begin{equation}
Q_{\mathrm{VLM}}(S)
=
\frac{
s_{\mathrm{lc}}(S)+
s_{\mathrm{spa}}(S)+
s_{\mathrm{fc}}(S)
}{30}.
\label{eq:verifier_visual}
\end{equation}
The judge uses fixed scoring anchors and returns
structured JSON outputs. We set the temperature to
$0$ and the maximum output length to $300$ tokens.
Candidates with failed judge requests are excluded
from acceptance until a successful retry is obtained.

\paragraph{Edit acceptance.}
The inference-time verification score is
\begin{equation}
\begin{aligned}
V(S)
={}&0.4Q_{\mathrm{PHY}}(S)\\
&+0.3Q_{\mathrm{SPA}}(S)
+0.3Q_{\mathrm{VLM}}(S).
\end{aligned}
\label{eq:verifier_weighted_score}
\end{equation}

At each repair round, the policy proposes a single
atomic decision. For a valid and executable edit,
the verifier computes the realized quality improvement
\begin{equation}
\Delta V_t
=
V(\widetilde{S}_{t+1})-V(S_t).
\label{eq:verifier_delta}
\end{equation}
The tentative edit is committed only if
$\Delta V_t>\epsilon$, where $\epsilon=0.01$.
Otherwise, the current scene is retained.
The maximum repair budget is $T_{\max}=6$ rounds. The acceptance rule ensures improvement in the
weighted verification score, but does not guarantee
monotonic improvement in every individual scene
quality metric. In particular, improvements in
collision avoidance or spatial consistency may
come at the expense of support or accessibility.
Conservative repair therefore refers to
verification-gated local intervention rather than
strict preservation of all satisfied constraints.

\subsection{Conservative Closed-Loop Inference}
\label{sec:closed_loop_inference}
Algorithm~\ref{alg:scenepilot_inference_full} summarizes the final generator-agnostic inference loop. Each round proposes at most one atomic intervention. The candidate edit is committed only when it passes schema/capability validation, executes successfully, and satisfies the conservative verifier. The next observation is always reconstructed from the retained scene after acceptance or rollback.

\begin{algorithm}[h]
\caption{ScenePilot Conservative Closed-Loop Repair}
\label{alg:scenepilot_inference_full}
\begin{algorithmic}[1]
\Require Native scene $S_0^{(g)}$, adapter
$\mathcal{I}_g=(\phi_g,\psi_g,C_g)$, repair policy
$\pi_\theta$, maximum rounds $T_{\max}$,
acceptance threshold $\epsilon$
\Ensure Repaired native scene
\For{$t=0$ to $T_{\max}-1$}
    \State $z_t\leftarrow\phi_g(S_t^{(g)})$
    \State Render $I_t=\{I_t^{\mathrm{diag}},
    I_t^{\mathrm{ann\text{-}top}}\}$
    \State $o_t\leftarrow(I_t,z_t,C_g)$
    \State $\hat{a}_t\sim\pi_\theta(\cdot\mid o_t)$
    \If{$\hat{a}_t=\texttt{STOP}$}
        \State \Return $S_t^{(g)}$
    \EndIf
    \State $a_t\leftarrow\operatorname{ValidateAndProject}
    (\hat{a}_t;C_g,S_t^{(g)})$
    \If{$a_t$ is invalid}
        \State $S_{t+1}^{(g)}\leftarrow S_t^{(g)}$
        \State \textbf{continue}
    \EndIf
    \State $\widetilde{S}_{t+1}^{(g)}
    \leftarrow F_g(S_t^{(g)},\psi_g(a_t;S_t^{(g)}))$
    \State Evaluate realized improvement $\Delta V_t$
    \If{$\Delta V_t>\epsilon$}
        \State $S_{t+1}^{(g)}
        \leftarrow\widetilde{S}_{t+1}^{(g)}$
    \Else
        \State $S_{t+1}^{(g)}\leftarrow S_t^{(g)}$
    \EndIf
\EndFor
\State \Return $S_{T_{\max}}^{(g)}$
\end{algorithmic}
\end{algorithm}

\section{Additional Evaluation Details}
\label{app:sceneeval_details}

\subsection{SceneEval Aggregation}
\label{app:sceneeval_aggregation}

For compact cross-metric reporting, we compute Fidelity and Plausibility aggregates from the official SceneEval metrics. All constituent metrics are percentages in $[0,100]$.
\begin{equation}
S_{\mathrm{Fid}}
=
\frac{
\mathrm{CNT}
+\mathrm{ATR}
+\mathrm{OOR}
+\mathrm{OAR}
}{4},
\label{eq:sceneeval_fidelity_aggregate}
\end{equation}

Since $\mathrm{COL}_{\mathrm{ob}}$ and $\mathrm{OOB}$ are lower-is-better metrics, we complement them before aggregation so that all terms follow the same higher-is-better direction:
\begin{equation}
\begin{aligned}
S_{\mathrm{Plaus}}
&= \frac{
\begin{aligned}
&(100-\mathrm{COL}_{\mathrm{ob}})\\
&\quad + \mathrm{SUP} + \mathrm{NAV}\\
&\quad + \mathrm{ACC} + (100-\mathrm{OOB})
\end{aligned}
}{5},
\end{aligned}
\label{eq:sceneeval_plausibility_aggregate}
\end{equation}
We then report a balanced overall score:
\begin{equation}
S_{\mathrm{Overall}}
=
\sqrt{
S_{\mathrm{Fid}}
\,S_{\mathrm{Plaus}}
},
\label{eq:sceneeval_overall}
\end{equation}

The Overall score is a summary derived from official SceneEval outputs rather than a replacement for the original metrics. We retain all original SceneEval metrics in the main results and appendix tables.

\subsection{Perceptual Evaluation Protocol}
\label{app:perceptual_ranking}

For visual-semantic evaluation, each VLM judge independently scores every scene on \emph{layout coherence}, \emph{relation fidelity}, and \emph{object completeness}. For criterion $c$ and policy $m$, the independent score assigned to frozen scene $i$ is denoted by $s^{\mathrm{VLM}}_{i,m,c}$. We convert the scores obtained for the same scene into policy ranks and compute the average rank as
\begin{equation}
\bar r^{\mathrm{VLM}}_{m,c}
=
\frac{1}{N}
\sum_{i=1}^{N}
\operatorname{rank}_{m}
\left(
-s^{\mathrm{VLM}}_{i,m,c}
\right),
\label{eq:vlm_average_rank}
\end{equation}
where lower rank is better and tied scores receive average ranks.

For the human evaluation, participants are shown anonymized and randomly ordered outputs produced from the same frozen input and rank them according to the same three criteria. Let $r^{\mathrm{Human}}_{i,m,c,j}$ denote the rank assigned by participant $j$. We report the average human rank across scenes and valid raters:
\begin{equation}
\bar r^{\mathrm{Human}}_{m,c}
=
\frac{1}{N |\mathcal{J}|}
\sum_{i=1}^{N}
\sum_{j\in\mathcal{J}}
r^{\mathrm{Human}}_{i,m,c,j},
\label{eq:human_average_rank}
\end{equation}
where $\mathcal{J}$ denotes the set of valid human raters. This produces directly comparable aggregate rankings for the VLM and human evaluations while keeping the VLM judgments independent and blind.
\section{Prompt and Output Schemas}
\label{app:prompts}

\subsection{Prompt Audit and Frozen ScenePilot Policy Prompt}
\label{app:policy_prompt}

We audit prompts against the final protocol-clean SceneReverse materialization rather than historical training or demo artifacts. The active policy contract uses the five canonical edit operations \texttt{translate}, \texttt{rotate}, \texttt{scale}, \texttt{delete}, and \texttt{add}, with \texttt{STOP} as the only terminal decision. The final SFT materialization contains no generator identity, diagnostic/RAG/prior field, or free-form explanation.

The static text of the final SFT policy prompt is shown below. At materialization time, \texttt{SCENE\_JSON} and \texttt{HOST\_CAPABILITIES} are replaced by compact serialized JSON. The line breaks below are introduced only for readability; the runtime materializer concatenates the same text into a single prompt string.

\begin{tcolorbox}[
  title={Frozen ScenePilot policy prompt used for final SFT materialization},
  colback=gray!5,
  colframe=black,
  boxrule=0.6pt,
  sharp corners,
  breakable,
  lines before break=2,
  title after break={Frozen ScenePilot policy prompt (continued)}
]
\begin{lstlisting}[style=promptstyle]
You are a generator-agnostic ScenePilot policy.
Make exactly one atomic decision per round.
STOP is the only terminal control.
Use stable object IDs and obey host_capabilities.
Canonical actions are translate, rotate, scale, delete, add;
return exactly one JSON object and no explanation.
SCENE_JSON=<canonical current scene JSON>
HOST_CAPABILITIES=<backend capability JSON>
\end{lstlisting}
\end{tcolorbox}

\subsection{Multimodal Observation Contract}
\label{app:observation_prompt}

The final multimodal SFT record augments the text prompt above with two images bound to the same current state: an annotated top-down rendering and a diagonal rendering. Thus the policy observation is $o_t=(I_t,z_t,C_g)$, where $I_t$ contains the two current-state views, $z_t$ is the canonical scene JSON, and $C_g$ is the backend capability descriptor. The training prompt does not expose future states, target-action hints, oracle history, generator identity, or diagnostic labels. The same observation format is used during training and inference. After each repair round, the observation is reconstructed from the retained scene without explicitly conditioning on previous repair actions or execution history. The textual policy input consists of a fixed repair instruction and serialized scene and capability
information. The original scene-generation
instruction is used by the verifier rather than
provided as a separate input to the repair policy.

\begin{tcolorbox}[
  title={Capability-conditioned multimodal input template},
  colback=gray!5,
  colframe=black,
  boxrule=0.6pt,
  sharp corners,
  breakable,
  lines before break=2,
  title after break={Capability-conditioned multimodal input template (continued)}
]
\begin{lstlisting}[style=promptstyle]
Image 1: current annotated top-down view
Image 2: current diagonal view

You are a generator-agnostic ScenePilot policy.
Make exactly one atomic decision per round.
STOP is the only terminal control.
Use stable object IDs and obey host_capabilities.
Canonical actions are translate, rotate, scale, delete, add;
return exactly one JSON object and no explanation.

SCENE_JSON=<canonical current scene JSON>
HOST_CAPABILITIES={
  "translate": <true|false>,
  "rotate": <true|false>,
  "scale": <true|false>,
  "delete": <true|false>,
  "add": <true|false>,
  "STOP": true
}
\end{lstlisting}
\end{tcolorbox}

\subsection{Canonical Atomic Output Schema}
\label{app:canonical_schema}

Every policy round returns exactly one JSON object. The following schemas match the final canonical action interface; field names are part of the executable contract.

\begin{tcolorbox}[
  title={Canonical repair decision schema},
  colback=gray!5,
  colframe=black,
  boxrule=0.6pt,
  sharp corners,
  breakable,
  title after break={Canonical repair decision schema (continued)}
]
\begin{lstlisting}[style=promptstyle]
# STOP
{"decision":"STOP"}

# translate
{"object":"<stable_object_id>",
 "operation":{"type":"translate",
              "delta_position_m":[dx,dy,dz]}}

# rotate
{"object":"<stable_object_id>",
 "operation":{"type":"rotate",
              "delta_yaw_deg":degrees}}

# scale
{"object":"<stable_object_id>",
 "operation":{"type":"scale",
              "scale_factor":[fx,fy,fz]}}

# delete
{"object":"<stable_object_id>",
 "operation":{"type":"delete"}}

# add
{"object":"<new_stable_object_id>",
 "operation":{"type":"add",
              "category":"<category>",
              "reference_object_ids":["<existing_id>"],
              "position_m":[x,y,z],
              "yaw_deg":degrees,
              "physical_size_m":[sx,sy,sz]}}
\end{lstlisting}
\end{tcolorbox}
Existing edit targets and \texttt{add.reference\_object\_ids} must resolve to stable IDs visible in the current canonical state; \texttt{add.object} introduces the new stable ID. A backend may expose only a subset of the five edit operations. Unsupported proposals are rejected by \textsc{ValidateAndProject} rather than silently converted into another action. Legacy \texttt{move} is normalized to \texttt{translate} only at the compatibility boundary.

\subsection{Prompt Versioning and Excluded Historical Prompts}
\label{app:prompt_versioning}

Earlier development artifacts contain prompts for retrieval-augmented relation planning, functional-group generation, diagnostic reasoning, multi-action repair, and repeated-scale constraints. These prompts correspond to superseded Grow-and-Repair or diagnostic pipelines and are not used to define the final ScenePilot policy. We therefore omit them from the reproducibility specification instead of mixing historical interfaces with the final checkpoint protocol.

\subsection{GRPO VLM Reward Judge}
\label{app:grpo_vlm_judge}

During GRPO, we use GPT-5.4 as a VLM judge to
provide the visual-semantic component
$Q_{\mathrm{VLM}}$ of the realized-improvement reward.
The judge receives top-down and diagonal scene
renderings, together with the room type as auxiliary
textual context. We set the temperature to $0$ and
the maximum output length to $300$ tokens.

The judge evaluates each scene independently along three dimensions: \emph{layout correctness} ($lc$), which measures geometric and spatial validity, including collisions, boundary violations, spacing, circulation, alignment, orientation, and grouping; \emph{semantic plausibility} ($spa$), which measures whether object relationships, scales, orientations, and functional groupings form a believable indoor arrangement; and \emph{functional completeness} ($fc$), which measures whether the scene contains sufficient core furniture and functional structure to support its intended use. Each dimension receives a continuous score in $[0,10]$ together with a short justification grounded in visible evidence. The scalar visual-semantic quality score is defined as the arithmetic mean of the three dimensions:
\begin{equation}
    Q_{\mathrm{VLM}}(S)
    =
    \frac{
        s_{\mathrm{lc}}(S)
        +
        s_{\mathrm{spa}}(S)
        +
        s_{\mathrm{fc}}(S)
    }{3},
    \label{eq:vlm_reward_score}
\end{equation}
which corresponds directly to the \texttt{overall}
field returned by the judge. Unlike the normalized
inference-time score in
Eq.~\ref{eq:verifier_visual}, the GRPO reward uses
the original mean score in $[0,10]$.
During GRPO, we use the realized improvement
\begin{equation}
    \Delta Q_{\mathrm{VLM}}
    =
    Q_{\mathrm{VLM}}(\widetilde{S}_{t+1})
    -
    Q_{\mathrm{VLM}}(S_t),
    \label{eq:vlm_reward_delta}
\end{equation}
rather than the absolute scene score. Therefore, the VLM reward assigns credit to the visual-semantic consequence of the sampled intervention rather than to the pre-existing quality of the input scene. This reward-side judge is used only for GRPO policy optimization and is distinct from the independent VLM evaluator used for final test-time reporting.

For reproducibility, the complete frozen prompt used by the GRPO VLM reward judge is provided below.

\Needspace{16\baselineskip}
\begin{tcolorbox}[
  title={Frozen VLM judge prompt used for GRPO reward},
  colback=gray!5,
  colframe=black,
  boxrule=0.6pt,
  sharp corners,
  breakable,
  title after break={Frozen VLM judge prompt used for GRPO reward (continued)}
]
\begin{lstlisting}[style=promptstyle]
You are an expert reviewer for indoor 3D scene layout and furniture arrangement.

You will be given rendered images of the SAME room, typically including:
- a top view, which is most reliable for judging global spatial arrangement, circulation, alignment, and object overlap;
- a diagonal view, which is most reliable for judging realism, object scale, accessibility, and whether furniture placement looks usable in perspective.

You may also be given optional textual context, such as the room type or a natural-language design request. If textual context is provided, use it only as auxiliary information. The primary judgment must come from the rendered images.

Your task is to evaluate the scene on exactly three criteria:

1) lc: Layout correctness
2) spa: Semantic plausibility
3) fc: Functional completeness

General evaluation principles:
- Be generous .
- Judge only what is visible in the provided images and optional textual context.
- Do not assume hidden furniture, invisible functional areas, or unshown geometry.
- If the two views disagree, use both views together and penalize uncertainty or visible inconsistency.
- Penalize physical or spatial problems such as:
  - obvious object-object collisions or severe overlap,
  - furniture intersecting walls or being partly outside the room,
  - blocked circulation or inaccessible furniture,
  - awkward spacing, unnatural placement, or unusable arrangements,
  - severe misalignment between related furniture pieces,
  - implausible scale or proportion,
  - missing core furniture required for the room's apparent function.
- Prefer concise but informative reasons.
- Use the full score range from 7 to 10 when appropriate.
- Scores do not need to be integers; use floats.

Detailed scoring criteria:

A) lc: Layout correctness
Definition:
Evaluate the geometric and spatial quality of the arrangement itself, regardless of style preference.

Focus on:
- whether objects are placed in reasonable positions inside the room boundary;
- whether there are collisions, overlaps, wall intersections, or out-of-bound placements;
- whether there is clear and usable free space for movement;
- whether furniture alignment, orientation, spacing, and grouping are spatially coherent;
- whether the overall composition looks organized rather than chaotic or arbitrarily scattered.

High lc score (8-10):
- Objects are well placed, mostly collision-free, inside the room, and support clear circulation.
- Relative positions and orientations are coherent.
- Major furniture is arranged in a spatially sensible and usable way.

Medium lc score (6-7):
- Layout is partially reasonable but has noticeable spacing issues, weak alignment, mild obstruction, or questionable placement.

Low lc score (3-5):
- Serious collisions, blocked walkways, unusable access, severe boundary violations, or obviously broken placement.

B) spa: Semantic plausibility
Definition:
Evaluate whether the scene makes semantic sense as a believable room layout for its apparent room type and intended use.

Focus on:
- whether object relationships are semantically appropriate;
- whether furniture types and pairings make sense together;
- whether items are positioned in functionally meaningful relations (e.g., seating around a table, bedside furniture near a bed, desk and chair pairing, TV facing seating, storage placed sensibly);
- whether scales, orientations, and usage relationships look realistic;
- whether the room appears like a plausible human-designed interior rather than a random collection of objects.

High spa score (9-10):
- Furniture relationships are natural and believable.
- The room reads clearly as an intended functional space.
- Object placement supports typical human use patterns.

Medium spa score (7-8):
- Scene is somewhat believable but contains odd pairings, weak semantic grouping, or several unnatural relations.

Low spa score (5-6):
- Scene appears semantically confused, implausible, or obviously unrealistic for the room type.

C) fc: Functional completeness
Definition:
Evaluate whether the room contains enough of the key furniture and arrangement structure needed to support its intended function.

Focus on:
- whether essential furniture for the apparent room type is present;
- whether the scene supports the main activity of the room;
- whether the functional zones look complete rather than partial or under-specified;
- whether the room feels missing major items that would normally be necessary.

Examples:
- A bedroom should usually include a bed and enough supporting furniture to make the room feel usable.
- A living room should usually include core seating and a coherent social or media focus.
- A dining room should usually include a dining table and appropriate seating.
- A study/workspace should usually include a desk or work surface and seating.
These are examples only; judge based on what is visible and any provided textual context.

High fc score (8-10):
- The room includes the major furniture needed for its purpose and feels functionally usable and reasonably complete.

Medium fc score (5-7):
- Some core functionality is present, but important supporting furniture or functional structure is missing.

Low fc score (3-4):
- The room is missing essential furniture and does not adequately support its intended use.

Scoring instructions:
- Score each criterion independently on a continuous scale from 0.0 to 10.0.
- Start from the visible evidence, not from a default high score.
- 8.0-10.0: strong quality with no major visible problems.
- 6.0-7.9: usable or plausible, with noticeable but non-critical problems.
- 3.0-5.9: major problems that substantially limit quality or use.
- 0.0-2.9: severely broken, unusable, or missing the essential room function.
- Apply these bands separately to lc, spa, and fc. A clean-looking render must not compensate for a missing function or an implausible arrangement.
- Do not infer problems or furniture that cannot be established from the images.
- Use decimal scores when the evidence supports a value between anchors.

Reason instructions:
- For each criterion, provide a short reason grounded in visible evidence.
- Mention the most important positive or negative factors only.
- Avoid long explanations, speculation, or restating the rubric.

Output instructions:
Return ONLY one JSON object with exactly this schema:
{
  "lc": {"score": 0.0, "reason": ""},
  "spa": {"score": 0.0, "reason": ""},
  "fc": {"score": 0.0, "reason": ""},
  "overall": 0.0
}

Additional output constraints:
- Do not output markdown.
- Do not output code fences.
- Do not output any text before or after the JSON object.
- The "overall" field must be the arithmetic average of lc.score, spa.score, and fc.score.
- Ensure the JSON is valid and directly parseable.
\end{lstlisting}
\end{tcolorbox}

\subsection{VLM-based Evaluation Protocol}
\label{app:vlm_eval}

For VLM-based evaluation, each rendered scene is judged independently without access to the generator identity, paired counterpart, or generation history. The judge receives only the rendered image and the corresponding task specification. We evaluate three dimensions: \textit{layout}, \textit{relation}, and \textit{completeness}. The exact prompt is shown below.

\begin{tcolorbox}[
  title={Frozen VLM judge prompt used for final evaluation},
  colback=gray!5,
  colframe=black,
  boxrule=0.6pt,
  sharp corners,
  breakable,
  title after break={Frozen VLM judge prompt used for final evaluation (continued)}
]
\begin{lstlisting}[style=promptstyle]
You are a blind judge of one rendered 3D scene. Evaluate only the attached
image and the task specification. Do not guess the generation method or
compare this scene with other scenes.

Score each requested metric independently with an INTEGER from 1 to 10.
Higher is better. Use these bands consistently:

1-2: Severe failure; most relevant requirements are visibly unmet.
3-4: Poor; several major problems.
5-6: Mixed; the main intent is recognizable, but noticeable problems remain.
7-8: Good; most requirements are met, with limited problems.
9-10: Excellent; all or nearly all visible requirements are met.

Within each band, choose the lower score for more numerous or more serious
problems, and the higher score for fewer or less serious problems.

Metrics:
- layout: Physical and spatial plausibility, including collisions,
  room-boundary violations, spacing, and circulation.
- relation: Whether the explicitly specified positional and facing
  relationships are satisfied.
- completeness: Whether the requested object categories and quantities
  are present.

Do not score texture, lighting, or artistic style. Do not count an occluded
object as definitely missing. If the image provides insufficient evidence
for a metric, return null for its score and explain why.

For each scored metric, pass = (score >= its pass_threshold).
Return valid JSON only.
\end{lstlisting}
\end{tcolorbox}



\section{Additional Qualitative Results}
\label{app:additional_qualitative}

\begin{figure*}[htbp]
\centering
\includegraphics[width=0.98\textwidth]{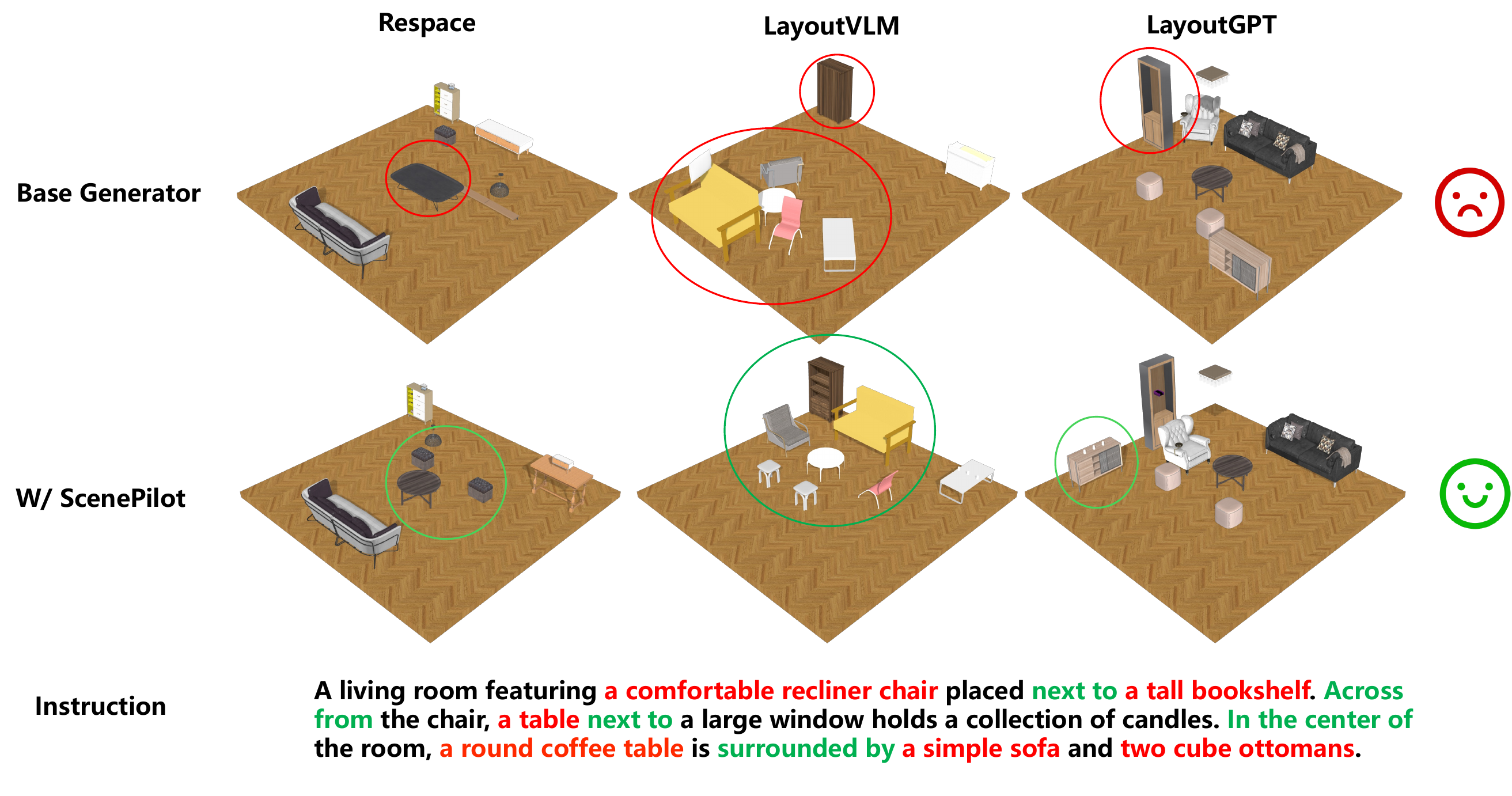}
\caption{\textbf{Additional qualitative results on a living-room scene.}
We compare the native outputs of ReSpace, LayoutVLM, and LayoutGPT
with the corresponding scenes repaired by the same frozen ScenePilot
policy. The highlighted regions show representative corrections to
object placement and prompt-specified spatial relations.}
\label{fig:additional_qualitative_living}
\vspace{-0.5em}
\end{figure*}

\begin{figure*}[htbp]
\centering
\includegraphics[width=0.98\textwidth]{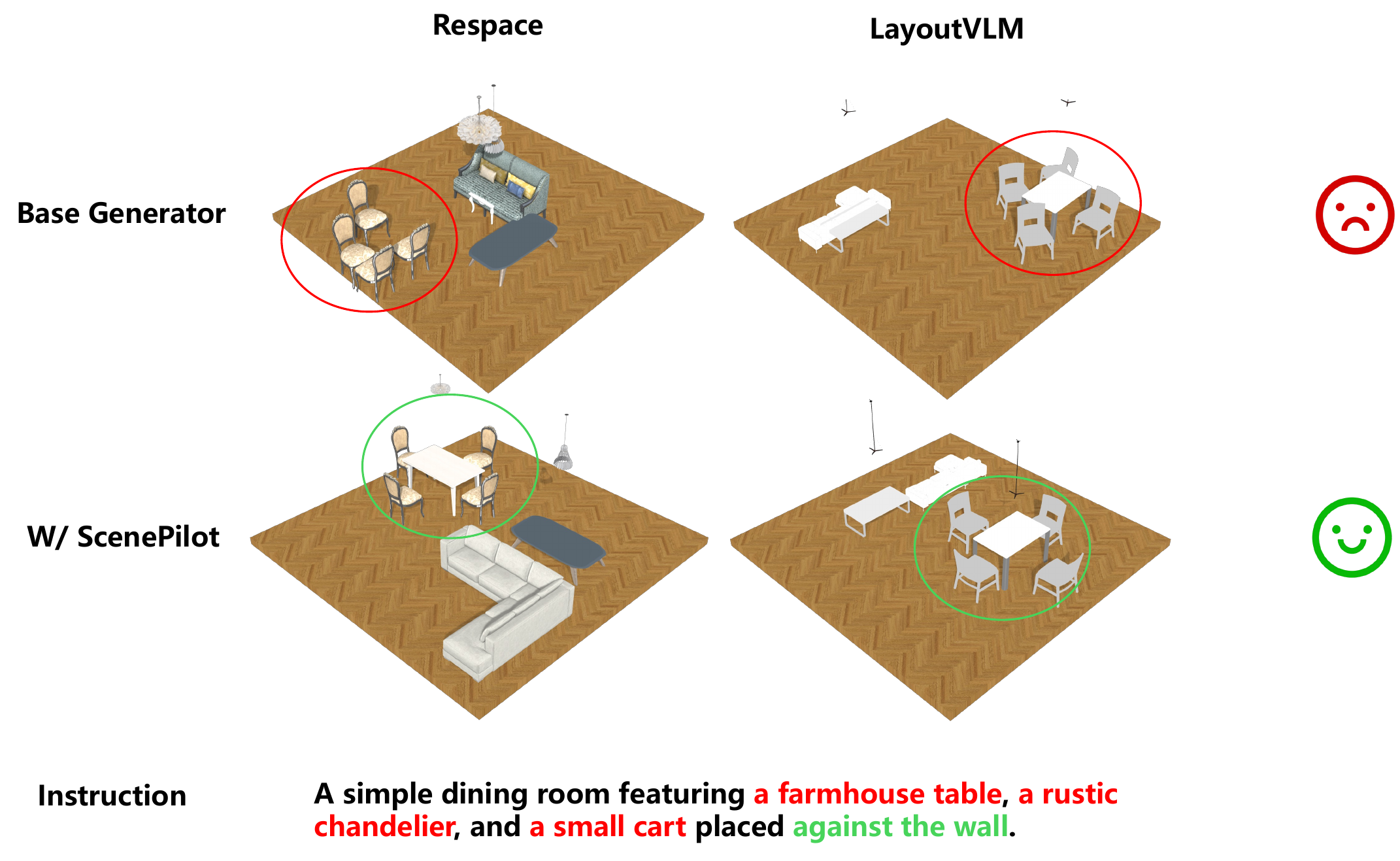}
\caption{\textbf{Additional qualitative results on a dining-room scene.}
We compare the native outputs of ReSpace and LayoutVLM with the
corresponding scenes repaired by the same frozen ScenePilot policy.
ScenePilot improves local object arrangement and spatial consistency
through localized edits without regenerating the scene from scratch.}
\label{fig:additional_qualitative_dining}
\vspace{-0.5em}
\end{figure*}

\begin{figure*}[htbp]
\centering
\includegraphics[width=0.98\textwidth]{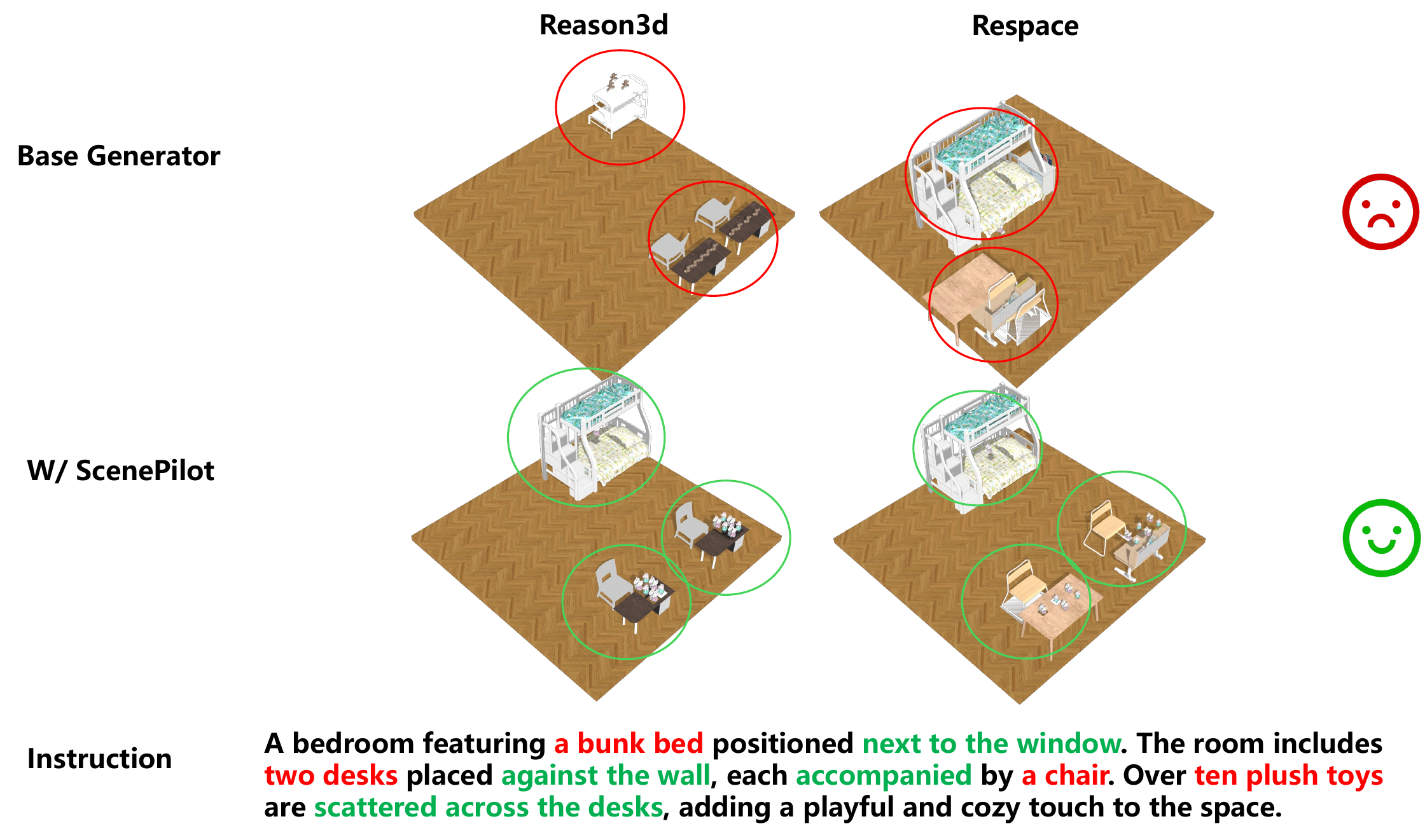}
\caption{\textbf{Additional qualitative results on a bedroom scene.}
We compare the native outputs of Reason3D and ReSpace with the
corresponding scenes repaired by the same frozen ScenePilot policy.
ScenePilot improves prompt-specified relations and object arrangement,
including the placement of the bunk bed, desks, chairs, and plush toys,
through localized scene edits.}
\label{fig:additional_qualitative_bedroom}
\vspace{-0.5em}
\end{figure*}

Figures~\ref{fig:additional_qualitative_living}--\ref{fig:additional_qualitative_bedroom} provide additional qualitative comparisons across different scene generators. For each example, we show the native generator output together with the scene repaired by the same frozen ScenePilot policy. Across generators, ScenePilot performs localized edits to improve object placement, orientation, spacing, and prompt-specified spatial relations while preserving the underlying generated scene. The highlighted regions illustrate representative corrections to geometric conflicts and relation inconsistencies.

\end{document}